\PassOptionsToPackage{table}{xcolor}
\documentclass{article} 
\usepackage{iclr2024_conference,times}

\usepackage{hyperref}       
\usepackage{url}            
\usepackage{microtype}            
\usepackage{graphicx}
\usepackage{subfigure}
\usepackage{booktabs}       
\usepackage{longtable}
\usepackage{lmodern}
\usepackage{rotating}
\usepackage{tabularx}
\usepackage{xltabular}
\usepackage{geometry}

\usepackage{titletoc}

\usepackage[utf8]{inputenc} 
\usepackage[T1]{fontenc}    
\usepackage{amsfonts}       
\usepackage{nicefrac}       
\usepackage{pifont}
\usepackage{bm}
\usepackage{wrapfig}
\usepackage{amsmath}
\usepackage{multirow}
\usepackage{marvosym} 

\usepackage{amssymb}
\usepackage{mathtools}

\usepackage[capitalize,noabbrev]{cleveref}

\title{MMIU: Multimodal Multi-image Understanding for Evaluating Large Vision-Language Models }

\iclrfinalcopy

\author{Fanqing Meng$^{*, 2, 1}$, Jin Wang$^{*, 3, 1}$, Chuanhao Li$^{*, 1}$,  Quanfeng Lu$^{1,2}$, Hao Tian$^{4}$\\ 
 \textbf{Jiaqi Liao}$^{1}$, \textbf{Xizhou Zhu}$^{5,1,4}$,   \textbf{Jifeng Dai}$^{5,1}$ \textbf{Yu Qiao}$^{1}$, \textbf{Ping Luo}$^{3,1}$, \textbf{Kaipeng Zhang}$^{1\dagger}$ \\
 \textbf{Wenqi Shao}$^{1\dagger}$\\\\
$^{1}$OpenGVLab, Shanghai AI Laboratory \quad $^2$Shanghai Jiao Tong University  \\
$^{3}$The University of Hong Kong \quad $^4$SenseTime Research \quad $^5$Tsinghua University \\
}

\begin{document}

\maketitle

\begin{center}
\vspace{-0.4in}
\textcolor{brown}{Project Page:}\,\,\href{https://mmiu-bench.github.io}{\color{brown}{https://mmiu-bench.github.io}}
\end{center}

\renewcommand{\thefootnote}{\fnsymbol{footnote}}
{\let\thefootnote\relax\footnotetext{
\noindent \hspace{-5mm}$\dagger$ Corresponding Authors: shaowenqi@pjlab.org.cn; zhangkaipeng@pjlab.org.cn \\
\noindent \hspace{-5mm}\quad \quad $*$ Equal contribution \\
}   }

\begin{abstract}
The capability to process multiple images is crucial for Large Vision-Language Models (LVLMs) to develop a more thorough and nuanced understanding of a scene. Recent multi-image LVLMs have begun to address this need. However, their evaluation has not kept pace with their development. To fill this gap, we introduce the Multimodal Multi-image Understanding (MMIU) benchmark, a comprehensive evaluation suite designed to assess LVLMs across a wide range of multi-image tasks. MMIU encompasses 7 types of multi-image relationships, 52 tasks, 77K images, and 11K meticulously curated multiple-choice questions, making it the most extensive benchmark of its kind. Our evaluation of 24 popular LVLMs, including both open-source and proprietary models, reveals significant challenges in multi-image comprehension, particularly in tasks involving spatial understanding. Even the most advanced models, such as GPT-4o, achieve only 55.7\% accuracy on MMIU. Through multi-faceted analytical experiments, we identify key performance gaps and limitations, providing valuable insights for future model and data improvements. We aim for MMIU to advance the frontier of LVLM research and development, moving us toward achieving sophisticated multimodal multi-image user interactions.
\end{abstract}

\section{Introduction}

The capability to process multiple images is crucial for multimodal large models, as a single image captures information from a specific angle and moment, limiting the model's ability to understand and reason about the entire scene \citep{song2024milebench,wang2024muirbench}. Multiple images, on the other hand, provide rich information from different perspectives and time points, enabling the model to synthesize this data and achieve a more comprehensive understanding, such as analyzing consecutive images for action prediction \citep{lu2024gui} or utilizing multi-view images in 3D navigation \citep{dai2017scannet}. The ability to process multiple images allows Large Vision-Language Models (LVLMs) to understand and handle complex visual tasks, thereby facilitating real-world applications.

\begin{figure*}
    \centering
    \includegraphics[width=0.9\linewidth]{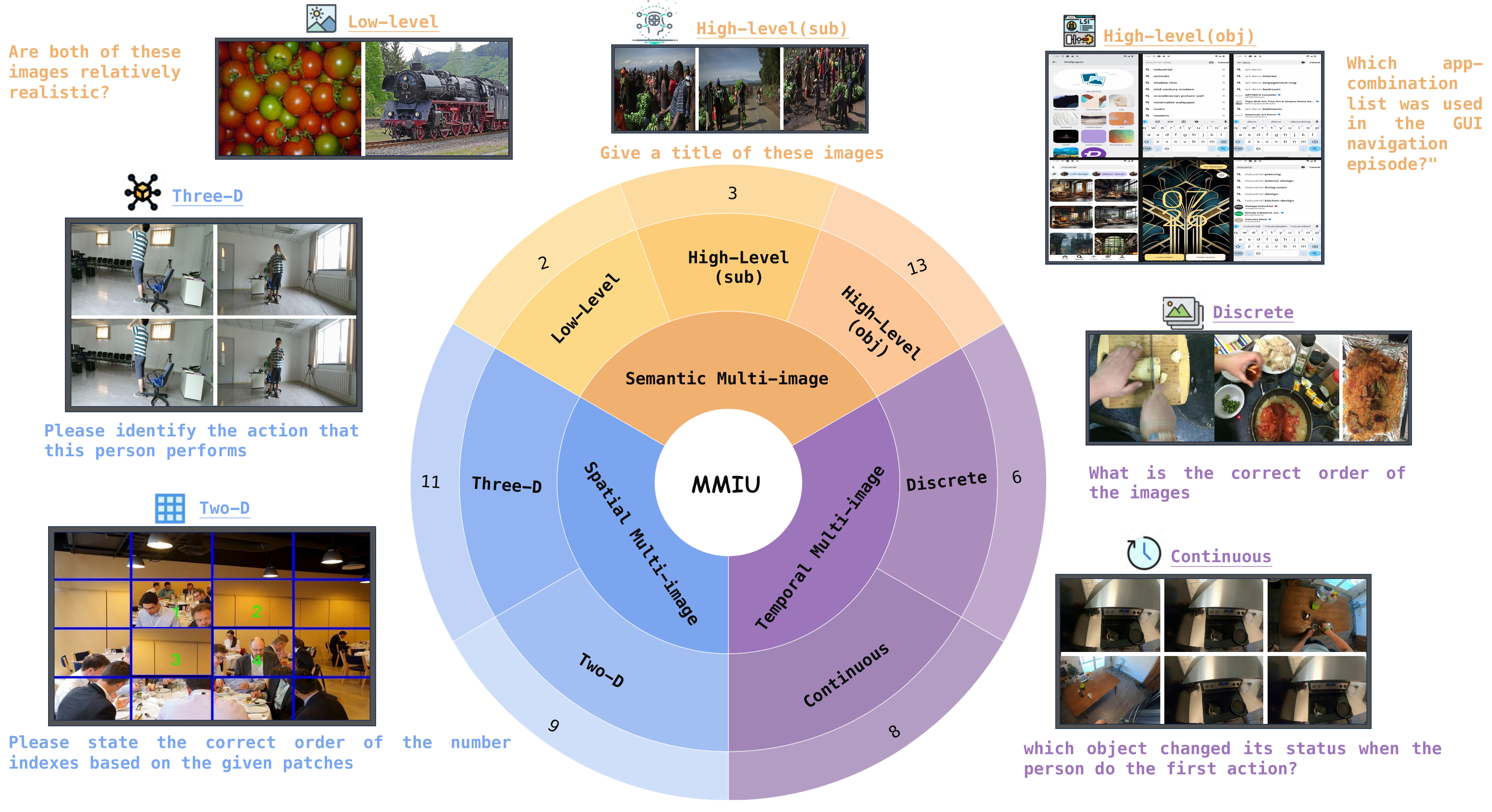}
     \vspace{-0.1in}
    \caption{Visualization of MMIU. Our MMIU contains 77,659 images, 7 types of image relationships, and 5 image modalities, along with 11,698 multiple-choice questions, providing a comprehensive evaluation for 52 multi-image understanding tasks. Each example comes from a task chosen from each multi-image relationship. We construct MMIU by adopting a top-down hierarchy where image relationships of interest are enumerated and multiple tasks are associated with each relationship. The number of tasks for each relationship is demoted.}
    \vspace{-0.2in}
    \label{fig:MMIU}
\end{figure*}

Due to the great importance of multi-image understanding, recent LVLMs have improved such a capability by pre-training on various image-text interleaved data such as M4-Instruct \citep{li2024llava}, Mantis-Instruct \citep{jiang2024mantis}, and OmniCorpus \citep{li2024omnicorpus}. However, the evaluation of multi-image LVLMs significantly lags behind their development. A good multi-image evaluation benchmark can help identify tasks that lead to poor performance and guide future model design data collection.
Prior datasets such as LVLM-eHub \citep{xu2023lvlm} and MMBench \citep{liu2023mmbench} focus on single-image tasks \citep{xu2023lvlm}, which cannot capture the complexity in multi-image scenarios. Although several recent benchmarks have attempted to evaluate the multi-image performance of LVLMs, they have limited coverage of multi-image tasks while capturing a few relationships between multiple images as shown in Table \ref{tab:comparison-MMIU}. For example, Video-MME \citep{fu2024video} focuses solely on temporal relationships and MUIRBENCH \citep{wang2024muirbench} does not consider spatial relationships between objects in multiple images, which is crucial in multi-image applications such as 3D navigation. Other works such as SlideQA \citep{tanaka2023slidevqa} and MMMU \citep{yue2024mmmu} focus on understanding and reasoning within specific input types or disciplines, preventing them from providing a general evaluation for multi-image capabilities. 



To build a comprehensive multi-image evaluation benchmark, we connect multi-image comprehension with manipulating information in working memory in cognitive psychology \citep{baddeley2000episodic}.
As pointed out by Multiple Trace Theory (MTT) \citep{moscovitch2006cognitive}, working memories are categorized into episodic memory which captures sequential information and can arrange events in the order they occur, semantic memory enabling concept comprehension, and spatial memory which helps understand spatial environments. 
Multiple images can be deemed as a visual memory. Understanding such a visual memory requires models to handle the semantic content, understand spatial relationships, and track temporal sequences of multiple images, closely mirroring human memory mechanisms. 
This inspires us to construct the evaluation benchmark to measure how well LVLMs tackle multi-image tasks from temporal, semantic and spatial perspectives.

This work introduces the Multimodal Multi-image Understanding (MMIU) benchmark, designed to comprehensively evaluate large visual language models (LVLMs) in multi-image task understanding. As shown in Table \ref{tab:comparison-MMIU}, we collect evaluation data through a top-down hierarchy, starting with the enumeration of image relationships spanning temporal, semantic, and spatial correspondences, and subsequently assigning multiple multi-image tasks to each relationship. The comprehensiveness of MMIU is twofold. First, it has the widest coverage of multi-image evaluation data to date, encompassing 7 types of multi-image relationships, 52 tasks (\emph{e.g.} multi-view action recognition), 77k images, and 11.6k carefully curated multi-choice questions, which is $1.81$ times larger than MilesBench \citep{song2024milebench}. Second, MMIU involves more diverse multi-image analysis tools than previous benchmarks, including performance comparison over image relationships, in- and out-of-domain task discovery by task map, and task learning difficulty by supervised fine-tuning (SFT). The multi-faceted analyses provide useful insights for model and data improvement.


We test 24 popular LVLMs on our MMIU, including closed-source models such as GPT4o \citep{gpt4o} and Gemini1.5 \citep{reid2024gemini}, and open-source models such as GLM4V  \citep{glm2024chatglm} and InternVL-Chat  \citep{chen2024internvl}. These LVLMs contain both multi-image (support multi-image input) and single-image (support only single-image input) models. For single-image models, we employ image concatenation to obtain the evaluation performance. 
The experimental results show that even the most advanced model, GPT4o \citep{gpt4o}, achieves only 55.7\% accuracy on MMIU, highlighting the inherent difficulty of these tasks. Other than the diverse analytical tools in Table \ref{tab:comparison-MMIU}, we conduct ablation studies to investigate the impact of unanswerable questions and multi-image concatenation methods on model performance. We summarize our findings as follows:


\begin{itemize}
    \item The best-performing model for multi-image tasks is GPT4o, with InternVL2 \citep{chen2024internvl} being the strongest among open-source models. The best closed-source model GPT4o leads the best open-source model InternVL2 by a large margin, (\emph{i.e.} 5.4\% accuracy). However, GPT4o achieves only 55.7\% accuracy on MMIU, indicating a substantial challenge in our benchmark.
     \item Some powerful LVLMs like InternVL1.5  \citep{chen2024internvl} and  GLM4V  \citep{glm2024chatglm} whose pre-training data do not contain multi-image content even outperform many multi-image models which undergo multi-image supervised fine-tuning (SFT), indicating the strong capacity in single-image understanding is the foundation of multi-image comprehension.
     
    \item By comparing performance at the level of image relationships, we conclude that LVLM excels at understanding semantic content in multi-image scenarios but has weaker performance in comprehending temporal and spatial relationships in multi-image contexts.
   
    \item The analysis based on the task map reveals that models perform better on high-level understanding tasks such as video captioning which are in-domain tasks, but struggle with 3D perception tasks such as 3D detection and temporal reasoning tasks such as image ordering which are out-of-domain tasks.

    \item By task learning difficulty analysis, tasks involving ordering, retrieval and massive images cannot be overfitted by simple SFT, suggesting that additional pre-training data or training techniques should be incorporated for improvement.

\end{itemize}

In summary, this paper makes three key contributions. First, we introduce and open-source the Multimodal Multi-image Understanding (MMIU) benchmark, a comprehensive evaluation suite that addresses various complex multi-image tasks, thereby filling a critical gap in multi-image comprehension. Second, our evaluation results demonstrate that current large visual language models (LVLMs), including proprietary models like GPT-4o, encounter significant challenges in solving multi-image tasks, particularly those involving spatial understanding. Third, we conduct multi-faceted analytical experiments, shedding light on the limitations and performance gaps of current models from various perspectives. We hope that MMIU will push the boundaries of LVLM research and development, bringing us closer to the realization of advanced multimodal multi-image user interactions.

\begin{table}[t!]
    \centering
    \small
    \caption{The comparison between MMIU and existing multi-image evaluation benchmarks including Video-MME  \citep{fu2024video}, MIRB  \citep{}, MUIRBENCH  \citep{wang2024muirbench}, and MileBench  \citep{song2024milebench}. We summarize the image relationships in previous benchmarks according to seven categories defined in Fig. \ref{fig:MMIU}. `Y\&N' indicates that our MMIU comprises both answerable and unanswerable questions. I, T, V, D and P represent image, text, video, depth map and point cloud, respectively. Compared with prior datasets, MMIU involves massive test samples spanning $52$ multimodal tasks and $5$ modalities, and comprehensive multi-image analyses by image relationships, task map and supervised fine-tuning (SFT). }
    \label{tab:comparison-MMIU}
    \scalebox{0.8}{%
        \begin{tabular}{c|cccccc| ccc}
            \toprule
            \multirow{2}{*}{Benchmark} &\multicolumn{6}{c|}{Data Statistics} &\multicolumn{3}{c}{Multi-image Analysis} \\
             \cmidrule{2-10}
            & \# Sample & \# Imgs.  & \# Relation & \# Task & \# Modality & Answerable? & Relation & Task Map & SFT\\
            \cmidrule{1-1}\cmidrule{2-10}

            Video-MME  & 2.7K &  - &    \textit{1}  &  \textit{30}  &  T,V & Y & - & \ding{55} & \ding{55}   \\
            MIRB  & 0.9K &  3.5k    &  3 &  11  & I,T,V  & Y & \ding{51} & \ding{55} & \ding{55}  \\
            MUIRBENCH   & 2.6K   &  11k     &  4 & 12 & I,T,V  & Y\&N & \ding{51} & \ding{55} & \ding{55}  \\
            MileBench   & 6.4K  &  97k   &  4  & 28  & I,T,V  & Y & \ding{51} & \ding{55} & \ding{55}  \\          
            \cmidrule{1-1}\cmidrule{2-10}
            \textbf{MMIU}  & 11.6K  &  77k   & 7 & 52 & I,T,V,P,D & Y\&N & \ding{51} & \ding{51} & \ding{51}  \\
            \bottomrule
        \end{tabular}%
    }
\end{table}

\section{Related Work}

\subsection{Large Vision-Language Models}

With the advancements in large language models (LLMs) \citep{touvron2023llama,jiang2024mixtral}, a series of studies have begun exploring multimodal LLMs capable of simultaneously interpreting visual and linguistic information. Through visual pre-training and instruction fine-tuning, LVLMs have demonstrated outstanding performance in understanding multimodal image-text inputs \citep{li2024llava,lu2024deepseek,bai2023qwen}. However, most LVLM training data consist primarily of single image-text pairs or pure text data, limiting their ability to comprehend multi-image inputs.
Therefore, researchers have considered using large-scale interleaved image-text corpora, such as MMC4 \citep{zhu2024multimodal} and Omnicorpus \citep{li2024omnicorpus}, during the pre-training phase of LVLMs. This approach has led to the development of models like Deepseek-VL \citep{lu2024deepseek} and Idefics \citep{laurenccon2024matters}, which exhibit notable performance in multi-image tasks. Building on this foundation, recent studies have applied instruction tuning with extensive multi-image data, resulting in models that handle multi-image tasks more effectively while utilizing fewer resources. Notable examples of these advancements include Mantis \citep{jiang2024mantis} and LLaVA-Next-interleave \citep{li2024llava}.
Nonetheless, the evaluation of these models' capabilities in handling multiple images has mainly been qualitative, and quantitative assessments of different models' performance across a broad range of multi-image tasks remain insufficiently explored.

\subsection{Large Vision-Language Models Benchmarks}

Benchmarking multimodal large language models (LVLMs) is crucial for identifying model limitations and guiding their development \citep{xu2023lvlm,ying2024mmt,liu2023mmbench}. Despite the existence of numerous benchmarks aimed at evaluating the perception or reasoning abilities of LVLMs, most of these benchmarks focus solely on single-image scenarios. 
Although some benchmarks include multi-image examples  \citep{jiang2024mantis,fu2024video}, they usually address limited capabilities. For instance, MANTIS-Eval \citep{jiang2024mantis} focuses on assessing a model's ability to perceive size, while Video-MME \citep{fu2024video} emphasizes image sequences and their temporal relationships. 
Recently, researchers have been dedicated to developing more holistic multi-image evaluation benchmarks, such as MileBench \citep{song2024milebench} and MUIRBench \citep{wang2024muirbench}, to provide a more thorough assessment of multi-image cognition. However, these benchmarks fall short in terms of task depth and breadth. For instance, MILEBENCH \citep{wang2024muirbench} provides a relatively comprehensive multi-image evaluation but lacks important multi-image tasks such as 3D spatial understanding and low-level semantics, which are essential for drawing complete conclusions. 
In contrast, MMIU offers a benchmark that combines both task depth and breadth, covering a wider range of image relationships, task types, and image categories. This enables a more comprehensive assessment of model capabilities.

\section{MMIU}

This section presents the proposed MMIU benchmark. MMIU is a comprehensive evaluation dataset encompassing 11K multi-choice questions for multi-image comprehension.
We first give a brief overview of MMIU in Section \ref{sec:mmiu-overview}. Then, we describe the construction process of MMIU in Section \ref{sec:mmiu-data-collection}.


\begin{wraptable}{rt!}{0.4\textwidth}
    \centering
    \caption{Key statistics for the MMIU}
    \scalebox{0.8}{
    \begin{tabular}{lr}
        \toprule
        \textbf{Statistic} & \textbf{Number} \\
        \midrule
        Total samples & 11698 \\
        Total images & 77659 \\
        Total tasks & 52 \\
        Img. relations & 7 \\
        Average images & 6.64 \\
        Average question words & 27.9 \\
        Range of images & 2$\sim$32 \\
        \midrule
        \textbf{Image Num Level} & \textbf{Number} \\
        \midrule

        - Few (2$\sim$5) & 7446 \\
        - Medium (6$\sim$15) & 2574 \\
        - Many (16$\sim$32) & 1666 \\
         \midrule
         \textbf{Unanswerable set} & \textbf{Percentage} \\
         \midrule
        - Replace keyword & 21\% \\
        - Replace answer image& 47\% \\
        - Replace other images & 53\% \\
        - Shuffle all images & 53\% \\
        - Irrelevant question/image set & 79\% \\

        \bottomrule
    \end{tabular}
    }
    
    \label{tab:dataset_statistics}
\end{wraptable}

\subsection{Benchmark Overview}\label{sec:mmiu-overview}

\begin{figure*}
    \centering
    \includegraphics[width=0.9\linewidth]{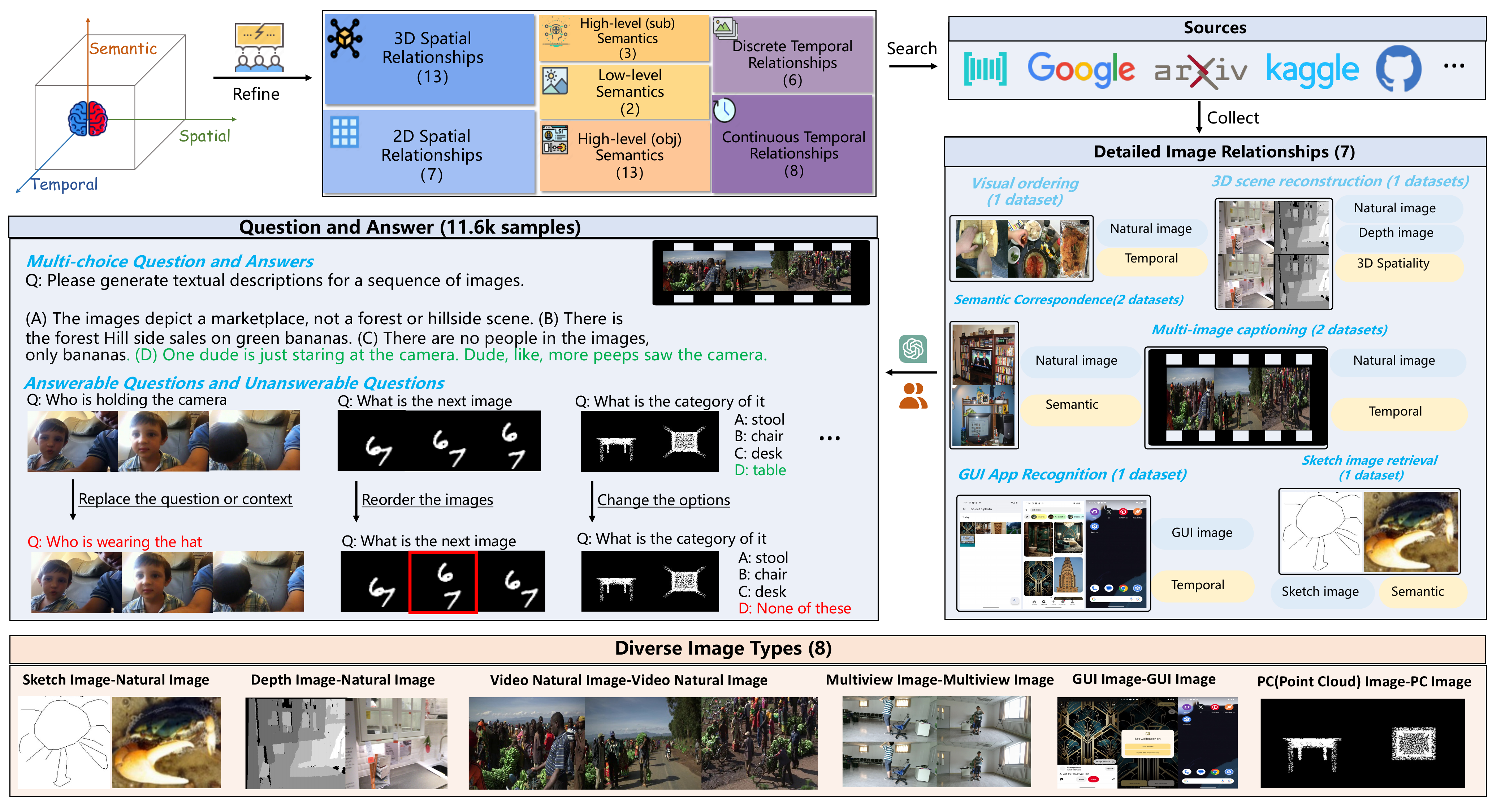}
    \vspace{-0.2in}
    \caption{An illustration of our data collection process. First, we refine multi-image tasks and collect task data based on cognitive psychology. Then, we standardize these datasets into a uniform format—metadata. Next, we generate multiple-choice samples with answerable and unanswerable questions from the metadata using either manually designed rules or GPT4o. Our benchmarks include capability evaluations across various image types.}
    \label{fig:construction}
\end{figure*}

MMIU is designed to measure multi-image understanding for LVLMs. It has two advantages compared with previous multi-image evaluation benchmarks as illustrated in Table \ref{tab:comparison-MMIU}. 
First, MMIU provides a comprehensive evaluation by encompassing massive test samples spanning various multi-image tasks and image relationships. Specifically, MMIU consists of 77,659 images and 11,698 multi-choice questions ($1.81$ times more than MileBench \citep{song2024milebench} which previously had the most multi-image test samples) with an average of 6.64 images per instance. It tests 7 distinctive multi-image relationships covering 52 diverse multi-image tasks, $1.73$ times more than VideoMME \citep{fu2024video} which previously contained the most multi-image tasks. In addition, we also create an unanswerable set comprising 19 tasks with each task containing 40 questions, considering that LVLMs cannot answer all questions in real scenarios. More detailed statistics of MMIU can be found in Table \ref{tab:dataset_statistics}. The diverse evaluation data requires the model to be capable enough to deeply understand semantical, temporal, and spatial clues in multi-images with various input types (Fig. \ref{fig:construction}).

Second, MMIU offers thorough analyses in multi-image understanding by utilizing multi-faceted analytical tools. 1) Thanks to the top-down hierarchy in collecting data, MMIU can compare performance across image relationships. 2) The extensive coverage of multi-image tasks enables evaluating on a task map, facilitating the discovery of in- and out-of-domain tasks. 3) The evaluation samples can be adapted to multi-image instruction tuning data. By SFT, the task learning difficulty can be acquired, which is crucial for the practitioner to improve the model and data.


\subsection{Data Curation Process}\label{sec:mmiu-data-collection}

Multi-image understanding is crucial for LVLMs, as multiple images are common media in real-world use. We treat a sequence of images as visual memories whose semantic, temporal, and spatial segments are crucial in retrieving information  \citep{moscovitch2006cognitive}. Following this inspiration, MMIU is built by collecting evaluation data through a top-down hierarchy, starting with the enumeration of image relationships spanning temporal, semantic, and spatial correspondences, and subsequently assigning multiple multi-image tasks to each relationship.

As shown in Figure \ref{fig:construction}, we first categorize multi-image relationships into semantic, spatial, and temporal relationships, which are further refined into seven basic types. Next, we collect data for each type of relationship and organize it into a standardized format. Finally, we construct multi-choice questions.

\textbf{Relationships $\rightarrow$ Tasks.} First, we divide the relationships among semantic, spatial, and temporal aspects in multiple images. For semantic relationships, we further refine them into \textit{1) Low-level semantic relationships} involving comparing low-level visual information features such as illumination, quality, and saturation. \textit{2) High-level (objective) relationships} among objects, attributes, and interactions between objects (\emph{e.g.}, a person hitting a ball, a person catching a ball). \textit{3) High-level (subjective) relationships} such as thematic associations, cultural connections, and emotional associations (\emph{e.g.}, the emotions expressed in these images). For temporal relationships, we refine them into \textit{4) Continuous temporal relationships} such as perception and inference tasks for video frame sequences. \textit{5) Discrete event sequence relationships}
such as understanding multi-step tutorials. For spatial relationships, we categorize them into \textit{6) 2D spatial relationships} such as rotation, translation, and symmetry. \textit{7) 3D spatial relationships} such as different camera perspectives and depth variations. The detailed information on each image relationship is shown in Section \ref{appendix:MMIU} of Appendix. Each image relationship is assigned several multi-image tasks whose correspondences are presented in Table \ref{tab:hierarchial} of Appendix.

\textbf{Tasks $\rightarrow$ Data. } We perform extensive searches for relevant datasets utilizing resources like Google, Paper With Code, and Kaggle, guided by the proposed tasks. Upon downloading the datasets, we thoroughly evaluate their appropriateness for the specific task, ensuring they are both usable and pertinent. 
%
We establish a standardized format, referred to as metadata, to organize the downloaded datasets. This format facilitates the creation of visual questions and answers. Each metadata includes a description of the task, as well as the question, answer, input context, and images for each sample. The detailed description of this format is in Table \ref{tab:metadata} in Appendix. We manually ensure the accuracy of this information and its convertibility into a multiple-choice question format. For efficient evaluation, each task is limited to a maximum of $200$ samples through random selection, aside from some tasks with insufficient data.


\textbf{Question and Answer Generation.} For each subtask, we create multiple-choice visual questions (with a maximum of eight options, depending on the task), with the choices and answers derived from their metadata. Specifically, depending on the task, we either manually design rules or use GPT4o  \citep{gpt4o} with carefully crafted prompts to ensure efficient and high-quality generation. For example, in 3D question-answering tasks, we instruct GPT4o to generate plausible but incorrect options based on the question and the correct answer. For image retrieval tasks, we randomly select incorrect images from the metadata as the wrong options. Additionally, we select 19 tasks and create 40 unanswerable samples for each task to construct an unanswerable set for robust evaluation. More details in unanswerable question generation are provided in Sec .\ref{appendix:construction}. 


\textbf{Challenges.} In constructing the MMIU, we encounter several challenges.
1) Designing plausible and accurate question templates. The designed questions should provide all the necessary information LVLMs may request, ensuring that they can derive the correct answer.
For example, in 3D object detection, each question should contain detailed camera pose information for the given images and specify the coordinate system where the detected objects are located.
2) Obtaining the correct answers with careful verifications. This is particularly challenging for tasks involving 3D spatial relationships.
For instance, in 3D pose estimation, the relative camera pose between images is not inherently provided in previous datasets  \citep{dai2017scannet}, which requires expert knowledge for accurate transformations.
Besides, examining the correctness of the obtained relative camera pose is also challenging, since they are more complex and abstract compared to the question answers regarding semantic/temporal relationships.
To tackle this, we transform the original camera pose of each individual scan to the relative camera pose required in MMIU through matrix multiplication.
Afterwards, we carefully examine the correctness of the obtained answer by applying the relative camera pose to image pairs, ensuring that the correspondences between images are correctly matched.
%
These challenges underscore the significant workload and difficulty involved in establishing MMIU as a comprehensive multi-image evaluation benchmark.


\begin{table*}[t!]
\centering
\caption{Quantitative results for 24 LVLMs across 52 tasks are summarized. Accuracy is the metric, and the Overall score is computed across all tasks.  The maximum value of each task is bolded. Notice that although InternVL1.5-chat supports multiple image inputs, its training phase did not incorporate multi-image data. The full term of task abbreviation can be found in Table \ref{tab:abbreviation} in Appendix.}
\label{tab:overall-results}
\resizebox{0.99\textwidth}{!}
{
\begin{tabular}{l|lllllllllllllllllllllllllll}
\toprule
Model & Overall & CR & ER & FD & FC & SC & VCor & VQA & VGR & FR & HR & I2IR & MIC & PR & S2IR & STD & STS & T2IR & VR & AQA & GAR & MVU & MEV & NIP & TL & TO & VidCap \\
  &   & GuAR & GNAP & TC & VClz & VCo & VO & EVQA & HE & IQASC & ICSC & ISTE & ITRSC & MAR & MR & JPS & 3DE & 3DOD & 3DOT & 3DPE & 3DSR & 3DQA & PT & RPM & SOT & 3DCR & 3DIR \\
\midrule
Frequency & 31.5 & 32.0 & 27.7 & 27.3 & 30.0 & 30.2 & 29.6 & 49.0 & 76.5 & 29.0 & 28.0 & 27.5 & 29.0 & 30.0 & 37.0 & 51.5 & 50.0 & 26.5 & 31.0 & 32.0 & 30.0 & 29.0 & 30.0 & 28.5 & 30.1 & 29.0 & 27.5 \\
  &   & 31.5 & 28.0 & 28.5 & 27.5 & 30.5 & 31.0 & 27.5 & 27.5 & 41.5 & 27.5 & 30.0 & 18.0 & 27.6 & 55.6 & 29.0 & 26.5 & 29.0 & 28.0 & 26.5 & 28.5 & 29.5 & 30.5 & 18.0 & 28.0 & 26.0 & 27.0 \\
\midrule
Random & 27.4 & 19.0 & 23.0 & 22.3 & 26.4 & 24.7 & 29.1 & 45.0 & 50.0 & 23.0 & 26.0 & 24.0 & 20.0 & 24.5 & 37.5 & 51.0 & 55.0 & 27.5 & 28.0 & 28.0 & 26.5 & 24.0 & 27.5 & 23.0 & 26.9 & 24.5 & 23.0 \\
  &   & 21.0 & 12.5 & 24.0 & 27.5 & 20.5 & 27.0 & 32.0 & 31.5 & 38.5 & 27.0 & 26.0 & 14.0 & 24.6 & 50.4 & 23.5 & 25.5 & 24.5 & 22.5 & 31.0 & 23.5 & 24.5 & 25.5 & 10.5 & 22.5 & 27.0 & 27.0 \\
\midrule
\rowcolor{lightgray} 
\multicolumn{28}{c}{\textbf{\textit{\large Closed-source LVLMs}}} \\ 
\midrule
GPT4o  & \textbf{55.7} &  67.8  & \textbf{46.5} & \textbf{88.8} & \textbf{42.6} & \textbf{41.5} & \textbf{72.6} &  79.2  &  61.3  &  76.0  &  42.0  &  59.5  & \textbf{93.5} &  61.5  &  67.0  &  11.0  & \textbf{84.0} & \textbf{70.5} &  68.0  &  33.5  & \textbf{91.5} &  71.5  & \textbf{35.0} &  26.5  & \textbf{50.8} & \textbf{28.0} & \textbf{92.5} \\
  &  &  78.0  &  46.5  & \textbf{62.5} & \textbf{43.5} & \textbf{97.5} &  21.5  &  57.5  & \textbf{29.5} &  88.0  &  58.5  & \textbf{35.0} &  17.5  & \textbf{81.9} &  46.6  &  23.5  &  24.0  &  40.5  & \textbf{94.5} &  85.0  &  22.0  &  39.0  &  55.0  &  12.5  &  56.0  & \textbf{69.0} & \textbf{49.0} \\
\midrule
Gemini1.5  &  53.4  &  71.0  &  31.8  &  73.5  &  24.3  &  34.9  &  47.3  &  78.8  &  61.0  &  88.0  & \textbf{80.0} &  74.0  &  89.0  &  70.5  & \textbf{81.5} &  74.0  &  80.0  &  60.5  &  68.0  & \textbf{35.5} &  88.0  & \textbf{75.0} &  25.0  &  21.0  &  45.6  &  26.5  &  84.0 \\
  &   & \textbf{93.0} &  39.5  &  59.0  &  30.0  &  60.0  & \textbf{43.5} &  53.5  &  22.5  & \textbf{91.0} & \textbf{64.5} &  24.0  &  13.0  &  68.8  &  51.1  & \textbf{34.5} &  20.0  &  32.0  &  48.5  &  37.5  &  28.5  &  35.5  &  66.5  &  13.0  &  61.0  &  55.0  &  43.0 \\
\midrule
Claude3.5  &  53.4  &  70.2  &  38.5  &  76.6  &  31.3  &  34.9  &  57.0  &  77.8  &  54.5  &  92.0  &  79.0  &  62.0  &  85.5  &  77.5  &  68.0  &  80.0  &  57.5  &  65.5  &  79.0  &  26.0  &  80.5  & \textbf{75.0} &  33.5  &  10.5  &  43.5  &  23.0  &  91.0 \\
  &   &  88.5  & \textbf{55.0} &  56.0  &  26.5  &  67.5  &  38.5  &  53.5  &  23.0  &  78.5  &  52.0  &  32.0  &  4.0  &  64.8  &  42.1  &  31.5  &  23.5  & \textbf{41.0} &  32.0  & \textbf{99.5} &  21.5  &  28.5  &  78.5  &  10.5  & \textbf{67.5} &  53.5  &  36.5 \\
\midrule
Gemini1.0  &  40.2  &  63.2  &  26.5  &  36.6  &  27.5  &  28.3  &  30.3  &  60.8  & \textbf{71.0} &  25.0  &  24.5  &  28.0  &  84.0  &  21.0  &  44.0  &  71.0  &  48.0  &  27.0  &  31.5  &  34.5  &  89.0  &  73.5  &  29.0  &  21.5  &  37.3  &  23.5  &  90.0 \\
  &   &  87.0  &  35.5  & \textbf{62.5} &  24.5  &  42.0  &  23.0  &  45.5  &  17.0  &  53.0  &  55.0  &  22.5  &  16.0  &  71.9  &  43.6  &  28.0  &  22.0  &  28.0  &  36.0  &  7.0  &  24.5  &  39.0  &  17.0  &  12.0  &  47.0  &  53.0  &  33.5 \\
\midrule
\rowcolor{lightgray} 
\multicolumn{28}{c}{\textbf{\textit{ Adequate Multi-Image SFT LVLMs}}} \\ 
\midrule
Mantis & 45.6 & 61.5 & 31.8 & 57.0 & 24.3 & 28.1 & 30.9 & 59.8 & 65.2 & 66.5 & 54.0 & 63.5 & 71.0 & 57.5 & 64.5 & 96.0 & 65.5 & 46.5 & 70.5 & 17.5 & 81.0 & 58.5 & 28.5 & 26.0 & 23.8 & 27.0 & 85.0 \\
  &   &  73.5  &  34.0  &  51.5  &  31.0  &  14.0  &  20.0  &  54.5  &  23.0  &  66.0  &  48.0  &  23.5  &  13.0  &  71.4  &  47.4  &  27.5  &  23.5  &  24.0  &  26.0  &  22.5  &  25.0  & \textbf{50.5} &  76.0  &  13.5  &  50.0  &  59.0  &  40.5 \\
\midrule
Llava-interleave & 32.4 & 29.5 & 24.8 & 26.3 & 23.2 & 26.4 & 25.1 & 48.8 & 49.8 & 23.5 & 25.0 & 28.0 & 57.0 & 21.5 & 33.0 & 63.5 & 54.5 & 25.0 & 26.0 & 24.0 & 27.0 & 49.5 & 29.0 & 23.0 & 25.4 & 27.5 & 32.5 \\
  &   &  43.0  &  34.0  &  49.0  &  29.5  &  32.0  &  26.0  &  30.0  &  21.5  &  42.0  &  47.5  &  22.5  &  14.0  &  23.6  &  32.3  &  17.5  & \textbf{28.5} &  23.0  &  17.5  &  3.0  &  31.0  &  36.0  & \textbf{79.0} &  15.0  &  60.5  &  34.5  &  42.5 \\
\midrule
\rowcolor{lightgray} 
\multicolumn{28}{c}{\textbf{\textit{\large Multi-Image input LVLMs}}} \\ 
\midrule
InternVL2  &  50.3  & \textbf{77.8} &  41.5  &  62.8  &  24.6  &  25.3  &  35.3  & \textbf{82.5} &  59.8  & \textbf{93.5} &  47.0  & \textbf{85.5} &  92.5  & \textbf{82.0} &  73.0  &  19.0  &  77.0  &  54.5  & \textbf{83.5} &  22.0  &  86.5  &  68.5  &  33.0  &  20.5  &  26.9  &  25.0  &  88.0 \\
  &   &  91.5  &  40.5  &  52.0  &  25.5  &  78.0  &  35.0  & \textbf{63.0} &  28.5  &  77.5  &  41.5  &  26.0  & \textbf{20.0} &  78.4  & \textbf{55.6} &  27.5  &  25.5  &  28.0  &  20.0  &  26.0  & \textbf{41.0} &  43.0  &  48.5  &  13.5  &  59.5  &  51.5  &  31.0 \\
\midrule
internvl1.5-chat & 37.4 & 63.7 & 31.0 & 22.6 & 20.3 & 16.3 & 28.3 & 63.2 & 38.5 & 21.0 & 28.0 & 26.5 & 82.5 & 20.5 & 31.5 & 6.0 & 45.5 & 26.5 & 29.5 & 29.5 & 85.0 & 65.0 & 32.0 & 23.5 & 29.0 & 18.5 & 89.0 \\
  &   & 90.5 & 35.5 & 56.5 & 23.5 & 31.0 & 24.5 & 53.0 & 26.0 & 40.0 & 49.0 & 25.5 & 15.5 & 59.3 & 43.6 & 19.5 & 22.5 & 23.5 & 15.0 & 33.5 & 28.0 & 39.0 & 71.0 & 9.5 & 46.5 & 50.5 & 39.5 \\
\midrule
idefics2-8b  &  27.8  &  28.0  &  25.8  &  26.4  &  26.7  &  24.6  &  28.6  &  58.5  &  30.8  &  3.5  &  9.5  &  4.0  &  82.0  &  5.0  &  27.5  & \textbf{98.5} &  70.5  &  12.5  &  7.0  &  16.0  &  24.5  &  12.0  &  19.0  &  23.5  &  22.3  &  18.0  &  19.5 \\
  &   &  23.5  &  22.5  &  21.0  &  26.5  &  21.5  &  22.5  &  14.5  &  21.5  &  31.0  &  50.5  &  25.5  &  13.5  &  15.1  & \textbf{55.6} &  27.5  &  26.0  &  21.5  &  9.0  &  21.5  &  23.0  &  11.5  &  61.0  & \textbf{18.0} &  52.5  &  44.5  &  40.5 \\
\midrule
deepseek-vl-7b  &  24.6  &  2.2  &  22.2  &  29.1  &  23.3  &  28.2  &  29.0  &  49.0  &  65.5  &  20.5  &  25.0  &  25.5  &  72.5  &  21.0  &  30.5  &  65.0  &  54.5  &  25.5  &  31.0  &  0.0  &  6.0  &  0.0  &  0.0  & \textbf{27.5} &  31.1  &  15.5  &  2.0 \\
  &   & 10.0 & 14.0 & 5.5 & 17.0 & 30.5 & 21.5 & 0.0 & 23.0 & 45.5 & 42.0 & 24.5 & 0.0 & 2.0 & 44.4 & 20.5 & 24.5 & 24.5 & 0.0 & 7.5 & 0.5 & 1.5 & 78.0 & 0.5 & 62.5 & 40.5 & 38.5 \\
\midrule
XComposer2-1.8b & 23.5 & 24.5 & 23.0 & 19.1 & 16.4 & 18.4 & 10.0 & 27.8 & 27.5 & 13.0 & 12.0 & 26.0 & 55.5 & 19.5 & 33.5 & 17.0 & 54.0 & 10.5 & 1.5 & 25.0 & 59.5 & 37.0 & 25.5 & 0.0 & 24.4 & 13.0 & 68.5 \\
  &   & 59.0 & 28.0 & 34.0 & 25.0 & 28.5 & 17.0 & 17.5 & 0.5 & 29.5 & 48.0 & 6.0 & 7.5 & 33.2 & 41.4 & 7.0 & 0.0 & 15.5 & 17.0 & 28.0 & 2.0 & 29.0 & 33.5 & 9.0 & 27.5 & 11.5 & 3.0 \\
\midrule
deepseek-vl-1.3b & 23.2 & 1.2 & 27.5 & 21.4 & 23.1 & 26.7 & 30.0 & 45.2 & 54.8 & 20.5 & 25.0 & 25.5 & 46.0 & 21.0 & 30.5 & 89.0 & 0.0 & 23.0 & 31.0 & 0.0 & 1.0 & 2.5 & 0.0 & 23.0 & 26.4 & 20.0 & 1.0 \\
  &  &  6.5  &  13.0  &  3.5  &  11.5  &  33.0  &  20.0  &  0.5  &  25.0  &  44.5  &  38.0  &  24.0  &  1.0  &  0.0  & \textbf{55.6} &  31.0  &  26.0  &  31.0  &  0.0  &  19.5  &  0.0  &  1.5  &  66.5  &  3.0  &  61.5  &  45.5  &  29.0 \\
\midrule
flamingov2 & 22.3 & 25.5 & 25.8 & 24.6 & 21.6 & 25.0 & 28.2 & 34.5 & 49.0 & 14.5 & 19.0 & 13.5 & 22.5 & 17.5 & 26.0 & 39.0 & 49.0 & 20.0 & 27.5 & 10.0 & 13.5 & 16.5 & 30.0 & 20.0 & 18.7 & 24.5 & 22.5 \\
  &   & 25.0 & 21.5 & 25.5 & 25.0 & 14.5 & 13.5 & 15.5 & 27.5 & 4.0 & 25.5 & 23.0 & 7.0 & 22.1 & 3.0 & 1.5 & 26.5 & 22.0 & 35.0 & 17.0 & 28.5 & 20.5 & 23.5 & 11.5 & 31.0 & 25.0 & 23.5 \\
\midrule
XComposer2 & 21.9 & 24.0 & 21.0 & 10.8 & 5.8 & 0.0 & 0.0 & 34.2 & 24.0 & 14.5 & 2.5 & 23.0 & 63.5 & 19.0 & 26.0 & 14.5 & 31.0 & 9.5 & 28.5 & 31.5 & 59.5 & 44.0 & 30.0 & 4.5 & 15.5 & 12.0 & 66.0 \\
  &   & 55.0 & 35.0 & 42.5 & 22.5 & 2.5 & 19.0 & 20.0 & 8.0 & 15.5 & 45.0 & 0.0 & 0.0 & 20.6 & 0.0 & 16.5 & 0.0 & 7.0 & 0.0 & 4.5 & 0.0 & 33.5 & 63.0 & 1.5 & 38.5 & 42.0 & 33.0 \\
\midrule
qwen-chat & 15.9 & 20.5 & 2.5 & 13.3 & 2.5 & 9.9 & 5.9 & 31.2 & 23.8 & 10.5 & 19.5 & 12.5 & 41.0 & 5.5 & 13.5 & 29.5 & 45.0 & 3.0 & 12.0 & 10.0 & 52.5 & 18.5 & 16.5 & 2.5 & 3.6 & 5.5 & 47.0 \\
  &   & 29.0 & 23.0 & 18.0 & 6.0 & 6.0 & 6.0 & 32.0 & 9.0 & 13.5 & 17.0 & 15.5 & 3.5 & 40.2 & 15.8 & 16.5 & 16.5 & 22.5 & 17.5 & 13.0 & 14.5 & 14.0 & 8.0 & 3.0 & 8.5 & 1.5 & 0.5 \\
\midrule
idefics-9b-instruct & 12.8 & 10.8 & 0.2 & 0.2 & 0.8 & 0.0 & 9.4 & 23.0 & 13.0 & 2.5 & 22.0 & 14.0 & 70.0 & 3.0 & 14.5 & 40.5 & 34.5 & 3.5 & 2.0 & 4.0 & 1.5 & 20.0 & 3.0 & 15.5 & 0.5 & 3.0 & 10.0 \\
  &   & 37.0 & 27.5 & 48.5 & 23.0 & 0.0 & 5.5 & 5.0 & 3.0 & 9.0 & 16.0 & 0.0 & 0.0 & 6.5 & 12.8 & 1.0 & 15.5 & 10.5 & 0.5 & 36.5 & 5.5 & 2.5 & 44.5 & 1.5 & 35.0 & 0.0 & 0.0 \\
\midrule
qwen-base & 5.2 & 9.2 & 0.5 & 5.7 & 5.8 & 0.5 & 1.0 & 5.0 & 4.5 & 0.0 & 1.0 & 0.0 & 20.5 & 0.0 & 2.5 & 1.0 & 43.0 & 1.0 & 0.0 & 0.0 & 4.5 & 8.5 & 0.5 & 0.0 & 0.0 & 0.0 & 7.5 \\
  &   & 24.5 & 8.0 & 29.5 & 5.0 & 5.5 & 6.5 & 2.0 & 2.0 & 8.5 & 11.5 & 0.0 & 0.0 & 0.5 & 5.3 & 0.0 & 0.5 & 7.0 & 0.0 & 21.5 & 0.0 & 5.5 & 2.5 & 0.0 & 0.5 & 0.0 & 0.0 \\  
\midrule
\rowcolor{lightgray} 
\multicolumn{28}{c}{\textbf{\textit{ Single-Image input LVLMs}}} \\ 
\midrule
glm-4v-9b & 27.0 & 32.8 & 16.0 & 31.8 & 8.7 & 9.0 & 4.7 & 59.0 & 55.8 & 31.0 & 7.5 & 19.5 & 82.0 & 23.5 & 24.5 & 81.0 & 67.0 & 25.0 & 30.0 & 7.0 & 59.5 & 53.5 & 10.5 & 5.0 & 25.9 & 10.0 & 76.0 \\
  &   & 55.5 & 19.0 & 34.0 & 5.0 & 11.5 & 14.5 & 26.0 & 11.5 & 35.5 & 41.5 & 16.0 & 6.5 & 25.1 & 29.3 & 9.0 & 14.0 & 14.5 & 7.0 & 0.5 & 5.5 & 27.0 & 35.0 & 7.5 & 26.0 & 48.5 & 23.5 \\
\midrule
llava-next-vicuna-7b & 22.2 & 22.2 & 9.2 & 11.0 & 9.1 & 7.7 & 10.5 & 37.0 & 23.2 & 7.0 & 16.5 & 8.0 & 66.0 & 5.0 & 23.5 & 88.0 & 42.5 & 13.0 & 14.5 & 5.5 & 51.0 & 42.5 & 9.5 & 10.0 & 17.1 & 6.5 & 66.0 \\
  &   & 50.5 & 14.5 & 38.0 & 9.0 & 9.5 & 8.5 & 31.0 & 5.0 & 28.5 & 27.0 & 8.5 & 5.0 & 22.6 & 29.3 & 6.5 & 4.0 & 4.0 & 6.0 & 8.0 & 9.5 & 32.5 & 72.0 & 1.0 & 38.0 & 42.0 & 25.0 \\
\midrule

MiniCPM-Llama3-V-2-5 & 21.6 & 41.1 & 11.8 & 13.2 & 8.7 & 5.0 & 11.3 & 47.8 & 38.5 & 7.0 & 3.0 & 6.5 & 77.0 & 7.5 & 18.5 & 41.5 & 41.5 & 10.0 & 5.0 & 0.5 & 70.5 & 51.0 & 13.5 & 4.5 & 17.6 & 5.0 & 83.5 \\
  &   & 46.0 & 24.5 & 26.0 & 4.5 & 20.5 & 12.0 & 43.0 & 0.0 & 25.0 & 44.5 & 0.0 & 1.5 & 34.2 & 38.3 & 6.0 & 8.5 & 5.5 & 9.5 & 20.0 & 4.5 & 24.5 & 14.5 & 0.5 & 22.0 & 32.5 & 15.0 \\
\midrule

llava-v1.5-7b & 19.2 & 14.1 & 4.2 & 13.7 & 5.8 & 1.9 & 6.9 & 27.3 & 35.0 & 6.5 & 12.5 & 12.5 & 53.0 & 10.0 & 25.5 & 66.5 & 43.0 & 19.0 & 3.5 & 2.5 & 23.5 & 36.5 & 12.0 & 16.5 & 6.7 & 7.0 & 28.0 \\
  &   & 24.5 & 17.5 & 40.0 & 15.0 & 21.5 & 4.0 & 26.0 & 7.5 & 26.5 & 17.5 & 5.0 & 4.5 & 25.6 & 27.1 & 8.5 & 8.0 & 4.0 & 6.0 & 6.0 & 14.5 & 29.5 & 66.0 & 2.0 & 35.0 & 34.5 & 28.5 \\
\midrule
sharegpt4v-7b & 18.5 & 16.4 & 5.0 & 10.8 & 6.2 & 9.0 & 2.7 & 34.2 & 28.5 & 4.5 & 10.5 & 3.5 & 57.0 & 4.0 & 12.5 & 55.5 & 44.5 & 13.5 & 5.0 & 5.0 & 26.0 & 38.0 & 14.0 & 15.5 & 10.9 & 6.0 & 25.0 \\
  &   & 26.5 & 19.0 & 42.0 & 7.5 & 14.0 & 7.5 & 31.5 & 7.0 & 29.0 & 18.0 & 5.0 & 1.5 & 28.1 & 23.3 & 9.5 & 3.0 & 7.0 & 6.0 & 2.0 & 8.0 & 27.5 & 65.5 & 0.0 & 44.0 & 36.5 & 31.0 \\
\midrule
sharecaptioner & 16.1 & 20.7 & 22.2 & 27.2 & 10.2 & 9.1 & 21.0 & 39.5 & 37.0 & 7.0 & 5.0 & 6.0 & 47.0 & 5.0 & 17.0 & 25.0 & 35.5 & 12.5 & 13.0 & 5.5 & 14.5 & 4.5 & 3.0 & 6.0 & 18.1 & 5.5 & 21.5 \\
  &   & 17.0 & 22.5 & 18.5 & 12.0 & 14.5 & 11.0 & 23.5 & 7.0 & 25.5 & 22.0 & 5.5 & 2.0 & 16.1 & 43.6 & 9.0 & 2.5 & 1.5 & 1.5 & 5.5 & 8.0 & 26.5 & 47.0 & 2.0 & 28.0 & 16.5 & 9.0 \\
\midrule

monkey-chat & 13.7 & 8.4 & 8.0 & 5.9 & 9.2 & 6.7 & 8.1 & 23.5 & 25.3 & 4.5 & 6.0 & 1.5 & 34.5 & 2.0 & 9.0 & 40.5 & 40.5 & 12.0 & 2.5 & 6.5 & 16.5 & 14.5 & 10.0 & 12.5 & 18.1 & 6.5 & 19.5 \\
  &   & 10.0 & 8.5 & 17.0 & 8.0 & 13.0 & 7.5 & 15.5 & 7.0 & 27.5 & 17.0 & 5.5 & 3.0 & 10.6 & 22.6 & 9.0 & 5.5 & 8.0 & 6.0 & 5.5 & 7.5 & 34.5 & 51.0 & 1.5 & 17.0 & 36.0 & 8.5 \\ \bottomrule
\end{tabular}%
}
\end{table*}

\section{Experiment}

This section first introduces the experimental setup in Sec .\ref{sec:setup}, including the testing methods and models used. Following this, we present the main results and multi-faceted analyses in Section \ref{sec:res} and Section \ref{sec:multitask}, respectively. Ablation studies are included in Section \ref{sec:abli}. We put more detailed information and error cases analysis in the Section \ref{appendix:experiment} in the Appendix.

\subsection{Experiment Setup}
\label{sec:setup}

\textbf{LVLM Models. } Specifically, we select four closed-source models: GPT4o \citep{gpt4o}, Claude3.5-Sonnet \citep{Claude2023}, Gemini1.5 Flash Pro \citep{reid2024gemini}, and Gemini1.0 Pro Vision \citep{team2023gemini}. Additionally, we evaluate eleven open-source models that support multiple image inputs: Mantis \citep{jiang2024mantis}, InternVL2 \citep{chen2024internvl}, LLaVa-Next-Interleave \citep{li2024llava}, InternVL1.5-Chat \citep{chen2024far}, Qwen-Chat \citep{bai2023qwen}, Qwen-Base \citep{bai2023qwen}, Idefics-9B-Instruct \citep{laurenccon2024obelics}, FlamingoV2 \citep{awadalla2023openflamingo}, DeepSeek-VL-1.3B \citep{lu2024deepseek}, XComposer2-1.8B \citep{dong2024internlm}, DeepSeek-VL-7B \citep{lu2024deepseek}, Idefics2-8B \citep{laurenccon2024matters}, and XComposer2 \citep{dong2024internlm}. Furthermore, we include seven models that only support single image input including LLaVA-V1.5-7B \citep{liu2024improved}, Monkey-Chat \citep{li2024monkey}, ShareCaptioner \citep{chen2023sharegpt4v}, ShareGPT4V-7B \citep{chen2023sharegpt4v}, GLM-4V-9B \citep{glm2024chatglm}, LLaVA-Next-Vicuna-7B \citep{liu2024llava}, and MiniCPM-Llama3-V-2.5 \citep{hu2024minicpm}. The detailed description of these models can be found in Table \ref{tab:lvlm} in Appendix.

\textbf{Evaluation Method. } With OpenCompass \citep{2023opencompass}, we first match the model's response to the corresponding options. If a match cannot be made, we mark it as Z \citep{yue2023mmmu}. The accuracy is used as the metric. Specifically: 1) For cases where the input token is too long for the tested model, we randomly sample images until it can be tested. 2) For single-image models which tend to respond with the same option, we shuffle the original options and retest. A result is considered correct only if both tests yield the correct answer. 3) For closed-source models, if the model refuses to respond due to copyright issues with the images, we discard those samples. The detailed setup can be found in Sec \ref{appendix:prompts} in Appendix.

\subsection{Main Results}
\label{sec:res}

As shown in Table \ref{tab:overall-results}, we report the average accuracy of all models across all tasks alongside
{Random Choice} and {Frequent Choice} baselines, with "overall" representing the average accuracy on all tasks. Specifically, we have the following findings.

\textbf{Multi-image tasks present significant challenges.} GPT-4o leads all models but achieves an average accuracy of only 55.7\%. Other proprietary models, such as Gemini1.5 and Claude3.5-Sonnet, also score 53.4\%. Among open-source models, InternVL2 performs the best, surpassing the proprietary Gemini1.0 Pro Vision with an accuracy of 50.3\%. \textit{There is a substantial performance gap (5.4\% accuracy) between closed-source and open-source models in multi-image comprehension.} By contrast, open-source models like InternVL2 achieve comparable or even superior performance to closed-source models such as GPT-4o in benchmarks focused on single-image understanding \citep{yue2023mmmu,liu2023mmbench,ying2024mmt}.

\textbf{The strong capability in single-image understanding is the foundation of multi-image comprehension.} Several advanced models such as InternVL1.5 which have been trained with only single-image data can achieve good performance in MMIU. For instance, GLM4V reaches 37.4\% accuracy, surpassing multi-image models LLaVa-interleave and Idefics2. Such success stems from its powerful capability in single-image multimodal understanding. Besides, GLM-4V also outperforms many multi-image models such as DeepSeekVL. This is because GLM-4V supports an ultra-high resolution of 1120*1120, allowing it to understand concatenated images and to reason. For instance, in the video-captioning task, its accuracy reaches 76\%.

\textbf{Adequate multi-image supervised fine-tuning (SFT) can improve the performance of models on multi-image tasks.} Notably, we have observed that many models trained extensively with multi-image data during the pre-training phase did not achieve satisfactory results, such as idefics2 and Deepseek-VL. However, Mantis and LLaVA-interleave stand out among all models. Their common feature is extensive multi-image instruction fine-tuning during the SFT phase. For instance, although idefics2 is trained with a large amount of multi-image data during the pre-training phase, it is trained by a few multi-image data during the SFT phase. Mantis, after performing multi-image SFT on the basis of idefics2, achieved a 17.8\% accuracy improvement.

\begin{figure*}
    \centering
    \includegraphics[width=0.8\linewidth]{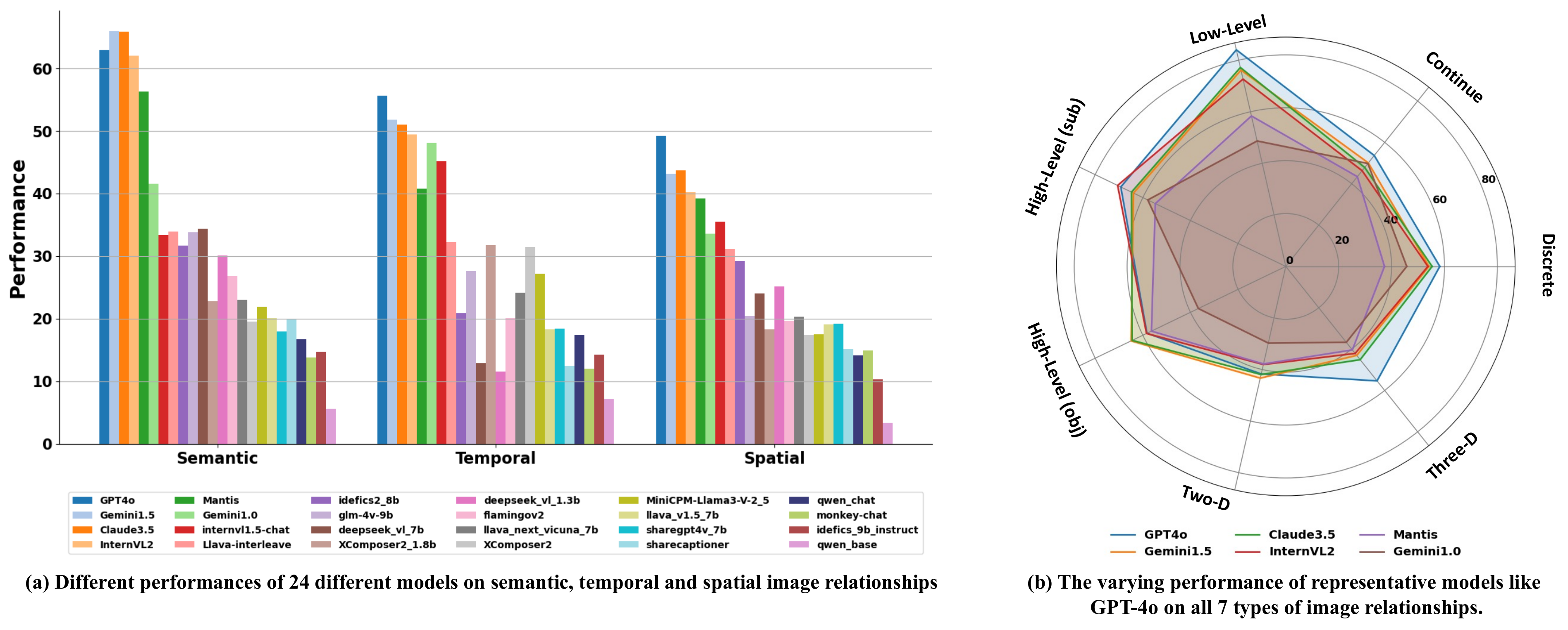}
    \caption{(a): The average performance comparison of 24 LVLMs on three main image relationships.
(b): The average performance comparison of representative models such as GPT4o on seven specific image relationships.}
    \label{fig:meta}
\end{figure*}

\subsection{Multitask Anaysis}
\label{sec:multitask}

\subsubsection{Performance across Image Relationships} 

As shown in Figure \ref{fig:meta}, models exhibit varying capabilities across different image relationships. More detailed visualizations can be found in Figure \ref{fig:meta_all} in the Appendix. In general,  LVLMs excel at understanding semantic content in multi-image scenarios, perform moderately in temporal tasks, and obtain the worst performance in
comprehending spatial relationships in multi-image contexts.

\textbf{1) In Semantic Relationships,} models generally perform well on multi-image semantic tasks involving low-level relationships. However, they struggle with high-level tasks, for subjective tasks such as Causality Reasoning and Emotion Recognition, which require the identification and reasoning of implicit visual information, highlighting a gap between model performance and human visual cognition. As for objective tasks such retrieval tasks, most models fail to tackle them. 
\textbf{2) In temporal relationships,} models can handle discrete and continuous temporal relationships relatively well but show mediocre performance on reasoning-intensive multi-image tasks. For instance, in sorting tasks, GPT4o achieves only 28\% and 21.5\% accuracy in temporal ordering and visual ordering tasks, respectively.
\textbf{3) In spatial relationships,} we find that models struggle with understanding both 2D and 3D positional relations. This is consistent with the observation in the previous single-image evaluation benchmark \cite{ying2024mmt} where they find that LVLMs fall short in localization and detection tasks requiring spatial reasoning. The tasks involving spatial relationships in MMIU become more challenging because models need to gather spatial information in multiple images and to reason.


\subsubsection{Analysis on the Task Map}

Task map is an effective tool for multi-task analysis \cite{ying2024mmt,ilharco2022taskvector}. Thanks to extensive coverage of multi-image tasks in MMIU, we build a task map to analyze the relationships between different tasks, allowing us to identify in- and out-of-domain tasks for current LVLMs.
%
Following MMT-Bench \cite{ying2024mmt}, we use QwenVL-chat to construct a task map where the distance between two tasks is given. Detailed construction process of the task map can be found in Sec .\ref{appendix:taskmap} in the Appendix.
%
In Fig. \ref{fig:taskmap} (a), we visualize the task map. After clustering through the task map, we visualize the model's performance on different clusters in Fig. \ref{fig:taskmap} (b) where task clusters are denoted by different colors.

\textbf{Tasks involving recognition or captioning are in-domain tasks} which can be handled by most current multimodal large models. For multi-image tasks, models generally struggle to achieve satisfactory results, obtaining good performance on a limited number of tasks. Specifically, for tasks in clusters 7, 8, and some tasks in cluster 2, which involve recognition or captioning (e.g., video captioning, action recognition), models perform relatively well. This is because these multi-image tasks focus on overall image perception, requiring less comparison and reasoning between images. 

\begin{figure*}
    \centering
    \includegraphics[width=0.9\linewidth]{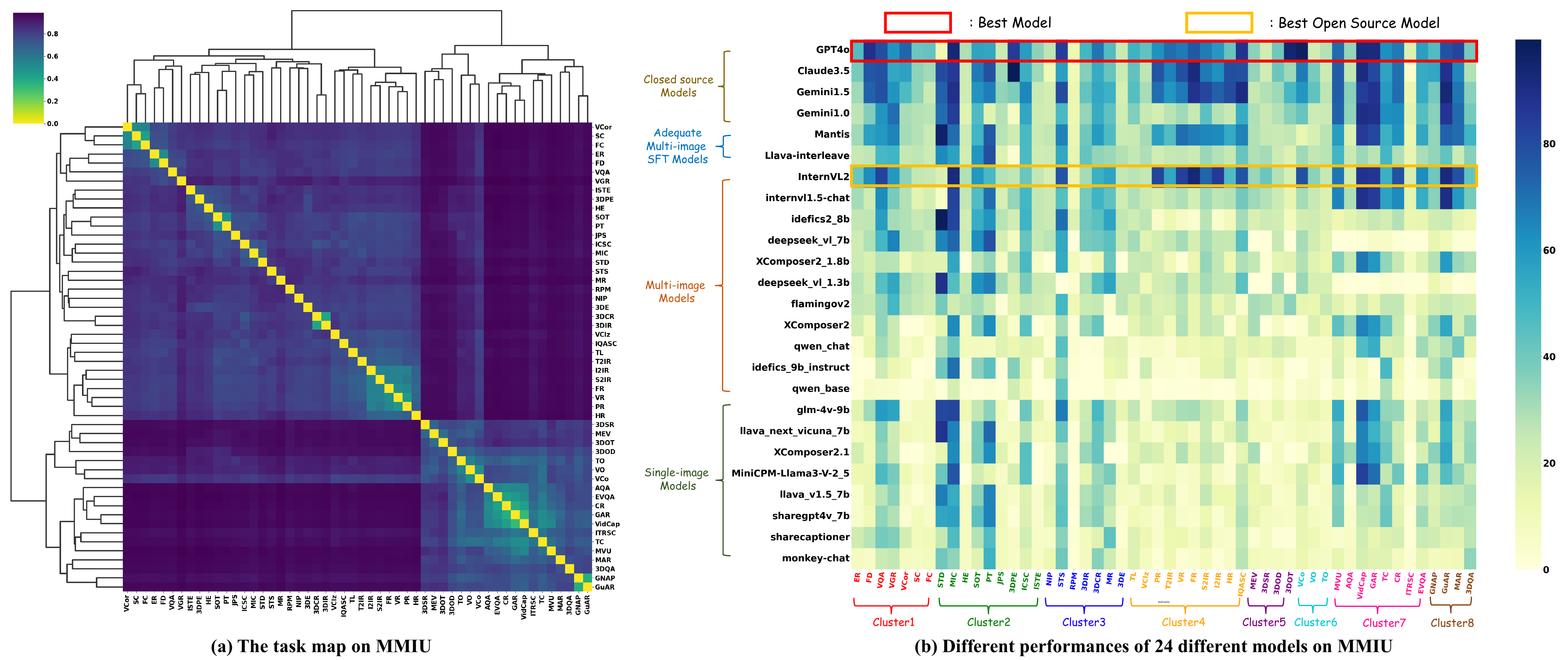}
    \caption{(a): Visualization of task maps and hierarchical clustering along with the task map. Please zoom in for clearer visualizations. (b): Visualization of model performance across various tasks. Different colors represent the respective categories formed through clustering, arranged sequentially from left to right, starting from the first category to the eighth. Notice that although InternVL1.5-chat supports multiple image inputs, its training phase did not incorporate multi-image data. }
    \label{fig:taskmap}
\end{figure*}

\textbf{Tasks involving temporal ordering and 3D spatial reasoning are out-of-domain Tasks} where most models perform poorly. Specifically, models struggle with tasks in clusters 4, 5, and 6. Clusters 4 and 6 involve modelling semantic relationships or sequential order among multiple images, requiring memorizing detailed long-context content and strong reasoning skills. Most LVLMs underperform on these tasks such as temporal ordering tasks).
Tasks in cluster 5 pertain to 3D visual tasks such as 3D detection and tracking. This may be due to the lack of 3D vision-language data in training LVLMs.



\subsubsection{Task Learning Difficulty}

\begin{wrapfigure}{r}{0.5\textwidth}
    \centering
    \includegraphics[width=0.9\linewidth]{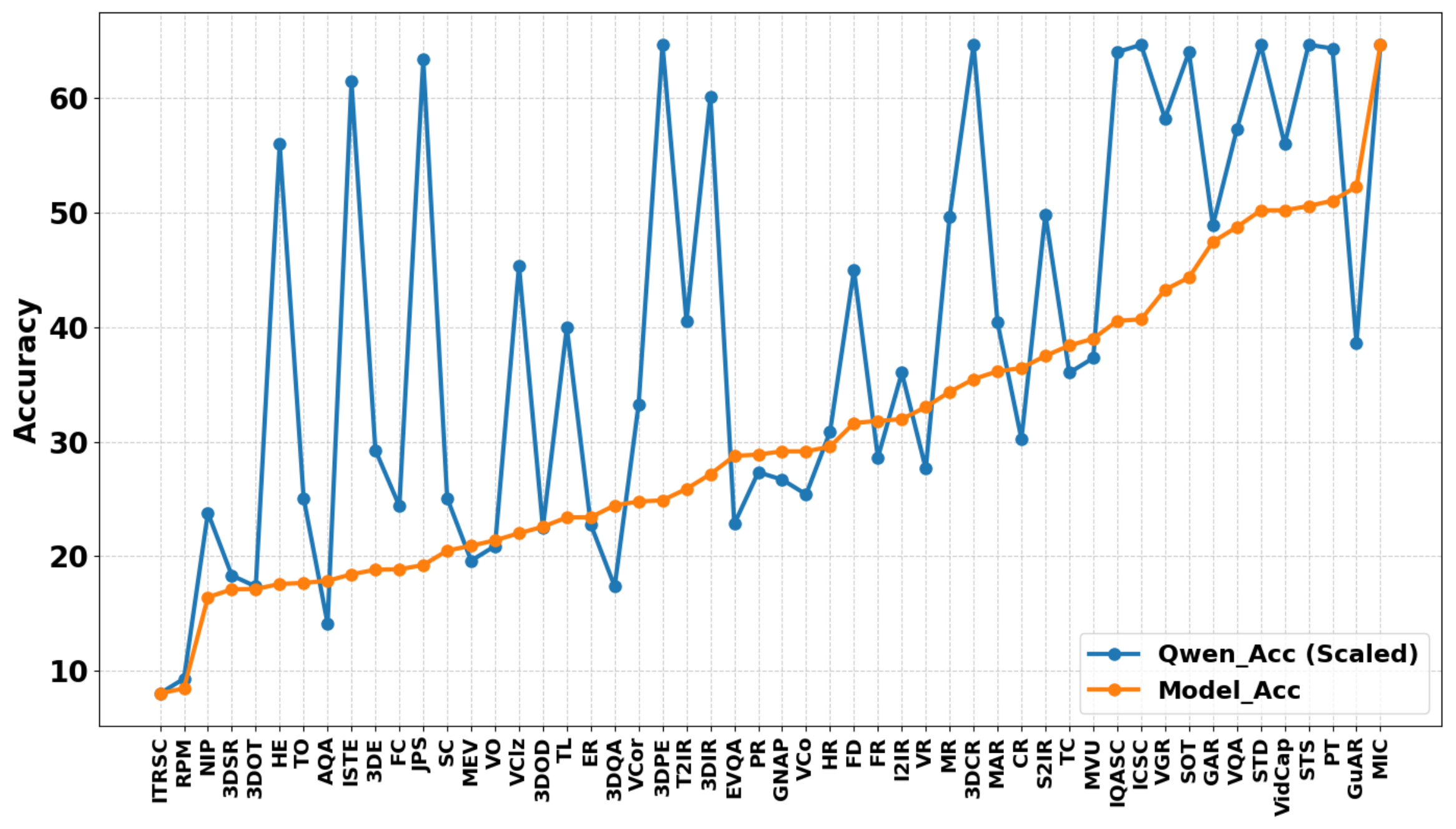}
    \caption{The performance of $Acc_{Model}$ and $Acc_{SFT}$ across different tasks, sorted by $Acc_{Model}$ in descending order, with $Acc_{SFT}$ scaled to the same magnitude as $Acc_{Model}$ for easy comparison.}
    \label{fig:Qwen_acc}
\end{wrapfigure}

We analyze task learning difficulty by SFT with all evaluation samples in MMIU being instruction tuning data. In this way, we can identify tasks which cannot be improved by simple SFT. To this end, we fine-tune QwenVL-chat on each task for 20 epochs and obtain the accuracy of QwenVL-chat on each task, denoted as $Acc_{SFT}$. The lower accuracy reflects the larger fitting difficulty of the task. Meanwhile, we also obtain the average accuracy of all tested models on each task, denoted as $Acc_{Model}$. This accuracy reflects the difficulty current models face in handling these tasks.

As shown in Figure \ref{fig:Qwen_acc}, we find that the Spearman correlation coefficient between $Acc_{SFT}$ and $Acc_{Model}$ is 0.66, indicating a high correlation. This suggests that both measures can reflect task difficulty to some extent. More importantly, we need to focus on tasks where both $Acc_{SFT}$ and $Acc_{Model}$ are low. A low $Acc_{SFT}$ indicates that the task is difficult to overfit even with SFT, suggesting that additional pre-training data or training techniques might be necessary. These tasks include 1) Ordering and retrieval tasks, which require strong memory and reasoning abilities—capabilities that are generally weak in large multimodal models.
2) Tasks involving a large number of images, such as EVQA, MEV, and GNAP, require models to support longer context lengths and possess strong memory capabilities. This indicates that future multimodal model designs should consider the ability to handle long contexts and emphasize the inclusion of multi-image data during the pre-training phase.


\subsection{Ablation Study}
\label{sec:abli}

\begin{figure*}
    \centering
    \includegraphics[width=0.9\linewidth]{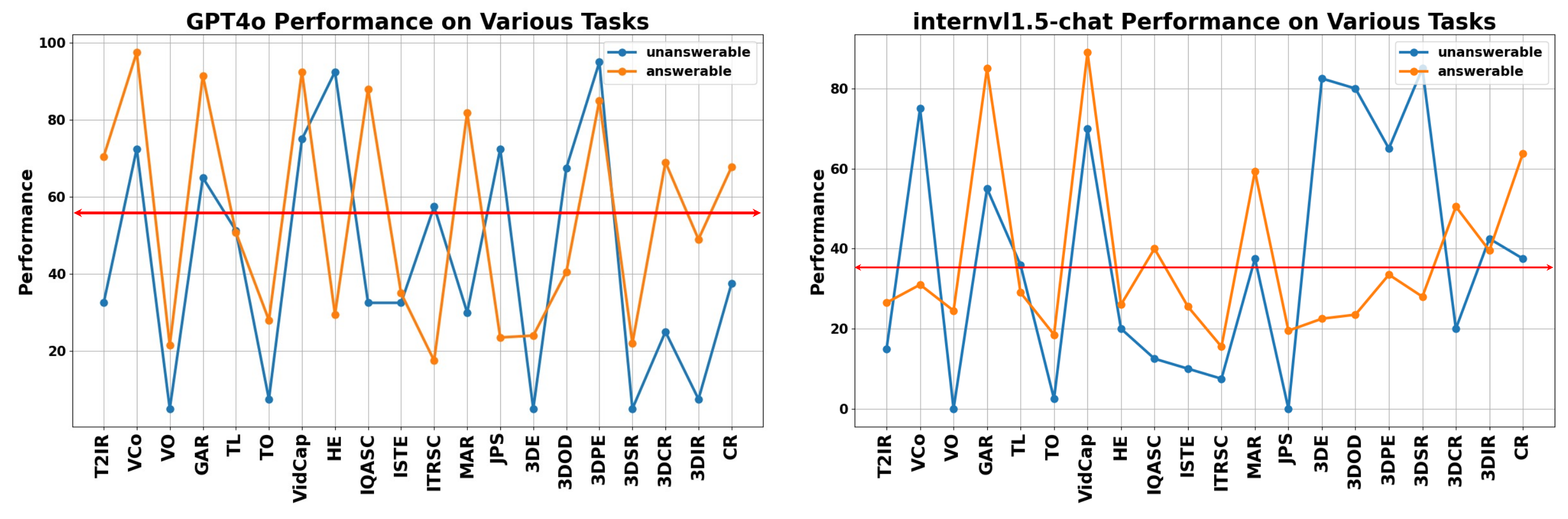}
    \caption{Comparison of GPT4o and InternVL1.5 on unanswerable and answerable questions, with the red line representing the model's average accuracy across all tasks.}
    \label{fig:unanswerable}
\end{figure*}

\textbf{Impact of Unanswerable Questions on Model Performance.} We have constructed 19 tasks, each including 40 questions. We tested a series of models on these questions, with full results referenced in Table \ref{tab:unanswerableall} in the Appendix. As shown in Figure \ref{fig:unanswerable}, we selected GPT-4o and InternVL1.5 as representative models for analysis. We observed that for some tasks where the models generally performed well, such as GAR (General Action Recognition), both GPT-4o and InternVL1.5 experienced performance degradation. However, for tasks that are inherently challenging for the models, as indicated by tasks below the red line in the figure, there is no significant pattern in the change of accuracy between answerable and unanswerable questions. We believe the reasons are as follows. 1) For tasks with high accuracy, introducing unanswerable questions confuses the models, increasing difficulty and thereby reducing accuracy. 2) For tasks with low accuracy, since the models already struggle with the original questions, the addition of unanswerable options might lead the models to directly choose the unanswerable option when uncertain, or the increased difficulty might further hinder their performance.

\begin{figure*}
    \centering
    \includegraphics[width=0.9\linewidth]{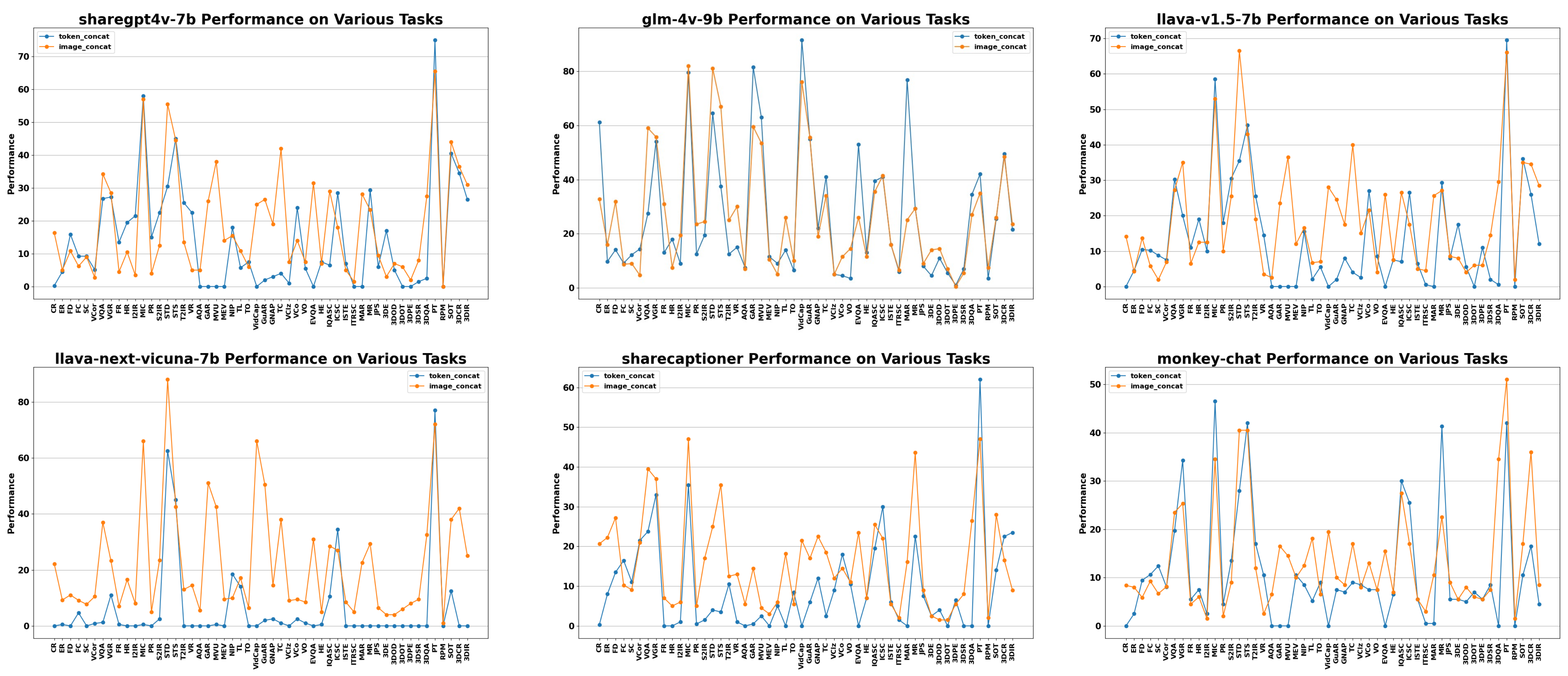}
    \caption{Comparison of the performance of different single-image models on various tasks in the MMIU when tested with image stitching or visual token stitching methods.}
    \label{fig:token-concat}
\end{figure*}

\textbf{Impact of Different Testing Methods on Model Performance.} 
For single-image input models handling multi-image tasks, one approach is to concatenate the images into a single image and feed it to the models. Besides, we explore an alternative method: concatenating all output visual embeddings before feeding them into LLMs. As shown in Figure \ref{fig:token-concat}, we observe that for these models, testing using concatenated visual tokens does not perform better than directly concatenating images. This is especially true for the LLavA series, where concatenating images significantly outperform concatenating visual tokens. In contrast, GLM-4V exhibits relatively consistent performance under both testing methods. 

\section{Conclusion}
In this paper, we present MMIU, a benchmark dedicated to comprehensively evaluating the performance of LVLMs on multi-image tasks. MMIU includes seven types of image relationships, such as 3D spatial relations, 52 tasks, and various image modalities, filling a gap in this field. We test 24 popular LVLMs on MMIU and analyzed the results using various analytical tools, including task maps. The experimental results indicate that current models, including GPT-4, struggle to handle complex multi-image tasks. We hope that MMIU will promote the development of more generalized capabilities in future models within the multi-image domain.

\bibliography{main}
\bibliographystyle{iclr2024_conference}
\clearpage
\newpage

\appendix

\section{MMIU Details} \label{appendix:MMIU}
\subsection{Multi-image Relations}
Overall, inspired by cognitive psychology, MMIU encompasses three broad types of image relationships: semantic, temporal, and spatial. Furthermore, we refine all detailed types as follows:

\begin{itemize}
    \item Low-level semantic relations: This mainly refers to multi-image comparisons of low-level visual features, such as lighting, quality, and saturation.

    \item High-level semantic relationship (objective): This refers to the objective assessment of high-level image features, such as objects (e.g., dog), attributes (e.g., number), and relationships between objects (e.g., person serving a ball, person catching a ball).

    \item High-level semantic relationship (subjective): This refers to the subjective assessment of high-level image features, such as thematic association (e.g., determining whether a set of images conveys a theme) or emotional association (e.g., identifying the emotions expressed in the images).

    \item Discrete time (event) temporal relationship: Compared to continuous video frames, this mainly refers to discrete event/time sequence image tasks, such as the analysis and reasoning of multi-step tutorials.

    \item Continuous time temporal relationship: It mainly refers to video frame sequence tasks, including perception (e.g., action classification) and reasoning (e.g., action prediction).

    \item Two-dimensional spatial relationship: This mainly refers to two-dimensional spatial multi-image relationships, such as rotation, translation, and symmetry.

    \item Three-dimensional spatial relationship: This mainly refers to multi-image relationships in three-dimensional spatial contexts, such as different perspectives and depth variations.
\end{itemize}

\subsection{Hierarchical Structure of MMIU}

\textbf{Image relationships and corresponding tasks.} We present all 7 types of image relationships in MMIU, totaling 52 tasks. Table \ref{tab:hierarchial} includes the distribution of tasks for each type of image relationship.

\begin{xltabular}{\textwidth}{>{\hsize=0.4\hsize}l|>{\hsize=1.6\hsize}X|>{\hsize=0.4\hsize}X}
\caption{Details of tasks classified by image relationship of our MMIU.} \label{tab:hierarchial} \\
\toprule \multicolumn{1}{c}{\textbf{Image Relationship}} & \multicolumn{1}{|c|}{\textbf{Task}} & \multicolumn{1}{c}{\textbf{\# Number}} \\ \midrule 
\endfirsthead

\multicolumn{3}{c}%
{\tablename\ \thetable{} -- continued from previous page} \\
\toprule \multicolumn{1}{c}{\textbf{Image Relationship}} & \multicolumn{1}{|c|}{\textbf{Task}} & \multicolumn{1}{c}{\textbf{\# Number}} \\ \midrule 
\endhead

\bottomrule
\endfoot

\bottomrule
\endlastfoot

Two-dimensional spatial relationship  & ravens-progressive-matrices, jigsaw-puzzle-solving, image-captioning-with-spatial-context, icon-question-answering-with-spatial-context, image-text-retrieval-with-spatial-context, image-spatial-transformation-estimation, homography-estimation, point-tracking, single-object-tracking & 9 \\
\midrule
Three-dimensional spatial relationship & threeD-scene-reconstruction, threeD-object-detection, egocentric-video-question-answering, threeD-object-tracking, threeD-pose-estimation, multiview-reasoning, multiview-action-recognition, threeD-depth-estimation, threeD-question-answering, threed-cad-recognition, threed-indoor-recognition &  11\\
 \midrule
Discrete time (event) temporal relationship & visual-coherence, textual-cloze, gui-app-recognition, gui-next-action-prediction, visual-cloze, visual-ordering & 6\\
 \midrule
Continuous time temporal relationship & general-action-recognition, video-captioning, next-img-prediction, temporal-ordering, meme-video-understanding, action-quality-assessment, temporal-localization, mevis
 & 8\\
 \midrule
Low-level semantic relations & visual-quality-assessment, forensic-detection & 2\\
\midrule
High-level semantic relationship (objective) & visually-grounded-reasoning, image2image-retrieval, sketch2image-retrieval, vehicle-retrieval, text2image-retrieval, face-retrieval, handwritten-retrieval, person-reid, spot-the-diff, spot-the-similarity, visual-correspondence, semantic-correspondence, functional-correspondence & 13 \\
\midrule
High-level semantic relationship (subjective) & emotion-recognition, casuality-reasoning, multiple-image-captioning & 3\\

 \bottomrule
\end{xltabular}

\textbf{Tasks and corresponding datasets.} To introduce MMIU more thoroughly, we need to introduce all 52 tasks in MMIU, their specific descriptions, and which datasets they come from. Table \ref{tab:temporaldataset}, Table \ref{tab:temporaldataset}, and Table \ref{tab:semanticdataset} respectively show the specific descriptions and data sources of tasks corresponding to temporal relationships, spatial relationships, and semantic relationships.

\begin{xltabular}{\textwidth}{>{\hsize=0.5\hsize}X|>{\hsize=1.5\hsize}X|>{\hsize=0.4\hsize}X}
\caption{Task descriptions and corresponding datasets for multi-image tasks in temporal relationships} \label{tab:temporaldataset} \\
\toprule \multicolumn{1}{c}{\textbf{Task Name}} & \multicolumn{1}{|c|}{\textbf{Task Description}} & \multicolumn{1}{c}{\textbf{Dataset}} \\ \midrule 
\endfirsthead

\multicolumn{3}{c}%
{\tablename\ \thetable{} -- continued from previous page} \\
\toprule \multicolumn{1}{c}{\textbf{Task Name}} & \multicolumn{1}{|c|}{\textbf{Task Description}} & \multicolumn{1}{c}{\textbf{Dataset}} \\ \midrule 
\endhead

\endfoot

\bottomrule
\endlastfoot

Action Quality Assessment & Action Quality Assessment involves evaluating the quality of an action or movement depicted in a sequence of natural images. Given a sequence of natural images capturing the action, the task requires assessing the quality of the action or movement. & Olympic \citep{parmar2017learning}, AQA-7 \citep{parmar2019action}\\ 
\midrule
Action Recognition & General Action Recognition is a vision task that involves recognizing and classifying the actions or activities depicted in a sequence of natural images. & Kinetics \citep{kay2017kinetics}\\ 
\midrule
Meme Video Understanding & Meme Video Understanding task involves understanding and interpreting the content and context of a meme video, where the visual input consists of a sequence of synthetic images. The task requires providing an explanation of the meme video content or context. & FunQA \citep{xie2023funqa}\\ 
\midrule
Mevis & MeVIS involves localizing objects of interest within a series of natural images. & MeVIS \citep{ding2023mevis}\\ 
\midrule
Next Image Prediction &  Next Image Prediction refers to predicting the image at the next moment based on a given series of images in chronological order.  & Moving MNist \citep{srivastava2015unsupervised}\\ 
\midrule
Temporal Localization &  Temporal Localization involves identifying the instance or target in a sequence of frames or a video at a specific time or time range. The task requires analyzing a sequence of natural images and determining the identifier of the target instance in the sequence.  & YouCook2 \citep{zhou2018weakly}, THUMOS14 \citep{wang2014action}\\ 
\midrule
Temporal Ordering & Temporal Ordering is a vision task that involves arranging a sequence of shuffled natural images in the correct temporal order.  &  Penn-Action \citep{chiu2019action} \\ 
\midrule
Video Captioning &  Video Captioning involves generating textual descriptions for a sequence of video frames, providing a narrative or informative explanation for the visual content.  & MSVD \citep{chen2011collecting}, MSRVTT \citep{xu2016msr}\\ 
\midrule
visual close &  Visual cloze style questions test a skill similar to that of textual cloze task with the difference that the missing information in this task reside in the visual domain   &  RecipeQA \citep{yagcioglu2018recipeqa} \\
\midrule
textual close &  Textual cloze style questions test the ability to infer missing text either in the title or in the step description by taking into account the question’s context which includes a set of illustrative images besides text   & RecipeQA \citep{yagcioglu2018recipeqa} \\
\midrule
visual coherence &  Visual coherence style questions test the capability to identify an incoherent image in an ordered set of images given the titles and descriptions of the corresponding recipe as the context.    & RecipeQA \citep{yagcioglu2018recipeqa} \\
\midrule
visual ordering &  Visual ordering questions test the ability of a system in finding a correctly ordered sequence given a jumbled set of representative images of a recipe. As in the previous visual tasks, the context of this task consists of the titles and descriptions of a recipe   & RecipeQA \citep{yagcioglu2018recipeqa} \\
\midrule
gui app recognition & Identify and analyze the applications utilized in the graphical user interface (GUI) segment of the episode.   &   GUI-Odyssey \citep{lu2024gui} \\
\midrule
gui next action prediction  & Predict the subsequent action based on the information provided in the previous screenshot and the given graphical user interface (GUI) navigation instructions.  & GUI-Odyssey \citep{lu2024gui} \\
\midrule

\end{xltabular}

\begin{xltabular}{\textwidth}{>{\hsize=0.5\hsize}X|>{\hsize=1.5\hsize}X|>{\hsize=0.4\hsize}X}
\caption{Task descriptions and corresponding datasets for multi-image tasks in spatial relationships} \label{tab:spatialdataset} \\
\toprule \multicolumn{1}{c}{\textbf{Task Name}} & \multicolumn{1}{|c|}{\textbf{Task Description}} & \multicolumn{1}{c}{\textbf{Dataset}} \\ \midrule 
\endfirsthead

\multicolumn{3}{c}%
{\tablename\ \thetable{} -- continued from previous page} \\
\toprule \multicolumn{1}{c}{\textbf{Task Name}} & \multicolumn{1}{|c|}{\textbf{Task Description}} & \multicolumn{1}{c}{\textbf{Dataset}} \\ \midrule 
\endhead

\endfoot

\bottomrule
\endlastfoot

Raven's Progressive Matrices & Raven's Progressive Matrices is a visual reasoning task involving synthetic images. Given a set of visual patterns, the task requires identifying the missing pattern from a set of options. & RAVEN \citep{zhang2019raven}, PGM \citep{santoro2018measuring} \\
\midrule
Jigsaw Puzzle Solving & Jigsaw Puzzle Solving task involves solving a jigsaw puzzle made up of natural images. The visual input consists of a shuffled patch of a natural image, and the instruction asks to rearrange the patches to reconstruct the original image. The patches can be fed as a set of images. & MSCOCO \citep{lin2014microsoft}, WikiArt \citep{saleh2015large} \\
\midrule
Image Spatial Transformation Estimation & Given pairs of images depicting scenes before and after a spatial transformation (e.g., rotation, translation), predict the type and magnitude of the transformation that occurred. & MSCOCO \citep{lin2014microsoft} \\
\midrule
Image Captioning with Spatial Context & Given a set of images (in NLVR, each sample can be split into 3 images), generate one sentence consistent with all images in terms of spatial context. & NLVR \citep{suhr2017corpus} \\
\midrule
Icon Question Answering with Spatial Context & Answer a multi-choice question in an icon image context. & IconQA \citep{lu2021iconqa} (a subset of it addresses spatial reasoning with multi-image.) \\
\midrule
Image Text Retrieval with Spatial Context & Given a text addressing spatial context, identify the matched image within candidates. & SPEC \citep{kocabas2021spec} \\
\midrule
Homography Estimation & Computing the 3x3 homography matrix that maps the coordinates of points in one image to their corresponding coordinates in another image. (Two images of the same planar.) & HPatch \citep{balntas2017hpatches}, Kaggle for HPatch \\
\midrule
Single Object Tracking & Visual Tracking involves following an object or region of interest across a series of images or frames. Given a query natural image with visual annotations, the task is to track the specified object or region in subsequent natural images. & TAP-Vid-DAVIS, TAP-Vid-RGB-stacking \citep{doersch2022tap} \\
\midrule
Point Tracking & Point Tracking involves locating and tracking a specific point of interest within a natural image. Given a query natural image with a visual mark indicating the initial position of the point, the task requires finding the same point within another natural image. & Mevis \citep{ding2023mevis} \\
\midrule
3D Classification - CAD & 3D classification - CAD involves classifying 3D images into specific categories based on their content and features. & ModelNet40 \citep{wu20153d} \\
\midrule
3D Classification - Indoor Point Cloud & 3D classification - indoor Point Cloud involves categorizing indoor scenes based on 3D point cloud data. & ScanObjectNN \citep{uy2019revisiting} \\
\midrule
Multi-view Reasoning & This task is centered on evaluating the multi-view reasoning capabilities of models. The objective is to deduce the relative camera motion based on two images of an object captured from different viewpoints. & BLINK \citep{fu2024blink} \\
\midrule
3D Object Detection and Pose Estimation & Detect objects and estimate their poses in 3D space using multiple views of the scene. Input Format: A Set of RGB images captured from different viewpoints, and a query image. Output Format: Detected objects with their 3D bounding boxes and poses based on the query image. & ScanNet \citep{dai2017scannet}, SceneNet \citep{handa2016scenenet}, SUN RGB-D \citep{song2015sun}, nuScenes \citep{caesar2020nuscenes} \\
\midrule
3D Scene Reconstruction & Reconstruct the 3D geometry of a scene. Input Format: An RGB image and a depth image. Output Format: A set of images captured from different viewpoints for this scene. & ScanNet \citep{dai2017scannet}, Matterport3D \citep{chang2017matterport3d}, SUN RGB-D \citep{song2015sun} \\
\midrule
3D Object Tracking & Input: Sequences of RGB-D images capturing object motion over time. Task: Track the movement of objects in 3D space across multiple frames. Output: Trajectories or paths of objects in 3D space (e.g., a sequence of 3D poses (position and orientation)). & KITTI \citep{geiger2013vision}, nuScenes \citep{caesar2020nuscenes} \\
\midrule
Multi-View Object Instance Segmentation & Estimate the instance-level segmentation map for a query image based on multiple images captured from different viewpoints. Input Format: A Set of RGB images captured from different viewpoints, and a query image. Output Format: A corresponding instance-level segmentation map for the query image. & ScanNet \citep{dai2017scannet}, SceneNet \citep{handa2016scenenet}, NYU Depth Dataset \citep{silberman2012indoor}, SUN RGB-D \citep{song2015sun} \\
\midrule
Multi-View Depth Estimation & Estimate the depth map for a query image based on multiple images captured from different viewpoints. Input Format: A Set of RGB images captured from different viewpoints, and a query image. Output Format: A corresponding depth map for the query image. & MegaDepth \citep{li2018megadepth}, SceneNet \citep{handa2016scenenet}, SUN RGB-D \citep{song2015sun} \\
\midrule
Multi-View Action Recognition & Recognize human actions or activities in a scene using information from multiple views. Input Format: A set of RGB images from multiple views. Output Format: Action labels/categories. & NTU RGB+D \citep{shahroudy2016ntu}, PKUMMD \citep{liu2017pku} \\
\midrule
3D Question Answering & Given inputs of the point cloud and a question about the 3D scene (real life), the model aims to output the correct answer. & ScanQA \citep{azuma2022scanqa}, NuScenes-QA \citep{qian2024nuscenes}, SQA3D \citep{ma2022sqa3d} \\
\midrule
Egocentric Video Question-Answering & Egocentric Video Question-Answering (EgoVQA) is a task that involves understanding and reasoning about activities and events from the first-person perspective. In this task, the model is presented with a sequence of egocentric (first-person) videos, typically captured by wearable cameras such as head-mounted cameras. The goal is to answer questions related to the content and context of the videos. & EgoTaskQA \citep{jia2022egotaskqa} \\
\midrule
Visual Navigation and Robotics & Given a series of images captured by robots or drones in different locations, the model outputs navigation commands or robot actions based on its spatial reasoning about the environment. Outputs may include directions for navigation, obstacle avoidance strategies, or object manipulation instructions. & DriveMLM (synthetic), YouTube-VIS \citep{yang2019video}, DAVIS \citep{pont20172017}, VOT2018 \citep{kristan2018sixth} \\
\midrule

\end{xltabular}

\begin{xltabular}{\textwidth}{>{\hsize=0.4\hsize}l|>{\hsize=1.6\hsize}X|>{\hsize=0.4\hsize}X}
\caption{Task descriptions and corresponding datasets for multi-image tasks in semantic relationships} \label{tab:semanticdataset} \\
\toprule \multicolumn{1}{c}{\textbf{Task Name}} & \multicolumn{1}{|c|}{\textbf{Task Description}} & \multicolumn{1}{c}{\textbf{Dataset}} \\ \midrule 
\endfirsthead

\multicolumn{3}{c}%
{\tablename\ \thetable{} -- continued from previous page} \\
\toprule \multicolumn{1}{c}{\textbf{Task Name}} & \multicolumn{1}{|c|}{\textbf{Task Description}} & \multicolumn{1}{c}{\textbf{Dataset}} \\ \midrule 
\endhead

\endfoot

\bottomrule
\endlastfoot

Visual Quality Assessment & This task is to evaluate the visual quality of two images, such as resolution, brightness, and clarity. & Q-bench \citep{wu2023q}, VE-LOL-L \citep{liu2021benchmarking}\\
\midrule
Forensic Detection & This task involves multiple images and requires determining which image is fake and not authentically composed. & FaceForensics++ \citep{rossler2019faceforensics++}, ForgeryNet \citep{he2021forgerynet} \\
\midrule
Visually Grounded Reasoning & This task involves giving a pair of images and checking if the sentence description matches the image pair. & NLVR v2 \citep{suhr2017corpus}, MaRVL \citep{liu-etal-2021-visually} \\
\midrule
Image-to-Image Retrieval & Image-to-Image Retrieval involves retrieving the candidate image ID that is most similar to the query image. & places365 \citep{zhou2017places}, tinyimagenet \citep{le2015tiny} \\
\midrule
Sketch-to-Image Retrieval & Sketch-to-Image Retrieval involves retrieving candidate images that are most similar to a given sketch image. & quickdraw \citep{DBLP:journals/corr/HaE17}, DomainNet \citep{peng2019moment} \\
\midrule
Text-to-Image Retrieval & Text-to-Image task involves generating an image based on a given textual description. The visual input consists of natural images, and the task instruction example could be 'Generate an image based on the provided text description.' The output provides the identifier of the generated image. & CUB220-2011 \citep{wah2011caltech}, Flowers102 \citep{nilsback2008automated}\\
\midrule
Person Re-Identification & Person Re-Identification involves identifying and matching a person's appearance across different camera views or over time. The task requires comparing a query image of a person with multiple candidate images to determine if the same person appears in the candidates. & Market-1501-v15 \citep{zheng2015scalable} \\
\midrule
Vehicle Re-Identification & Vehicle Re-Identification involves identifying a specific vehicle from a set of candidate vehicle images based on a given query image of the vehicle. & veri-776 \citep{liu2016deep} \\
\midrule
Face Verification & Face verification involves recognizing the identity of a query face image by comparing it with each support face image with an annotated identity. & LFW \citep{huang2008labeled}, CelebA \citep{liu2015deep} \\
\midrule
Handwritten Text Retrieval & Handwritten Text Retrieval and Verification involves retrieving and verifying handwritten text from a query image against candidate images containing handwritten text. & IAM \citep{marti2002iam} \\
\midrule
Spot the Difference & Spot the Difference task involves identifying the numeric value corresponding to the number of differences between two natural images. & spot-the-diff \citep{jhamtani2018learning} \\
\midrule
Spot the Similarity & Spot the Similarity involves identifying the similarity between multiple images and providing an explanation for the judgment. &  TLL \citep{rosenfeld2018totally}, DISC21 \citep{douze20212021} \\

\midrule
Visual Correspondence & This task involves providing several images from different angles and finding the same points in different perspectives, such as specific pixels. & BLINK \citep{fu2024blink}, ScanNet \citep{dai2017scannet} \\
\midrule
Semantic Correspondence & The task requires providing several images of different species and identifying semantically identical points across the different species, such as the head of a horse and the head of a human. & BLINK \citep{fu2024blink}, MISC210K \\
\midrule
Functional Correspondence & The task requires providing several images of different tools and identifying functionally identical points across the different tools, such as the handle of a broom and the handle of a toothbrush. & BLINK \citep{fu2024blink}, FunKPoint \\
\midrule
Emotion Recognition & The task is to provide multiple images, most of which depict the same emotion, and identify the one image that represents a different emotion. & FindingEmo \citep{mertens2024findingemo}, ExpW \citep{zhang2018facial} \\
\midrule
Casuality Reasoning & The task is to provide multiple images from a video and ask about the cause leading to a specific result, such as: "Why did the little girl stop the car?" The answer might be: "She stopped to wait for her mom." & NeXTQA \citep{xiao2021next}, VideoABC \citep{zhao2022videoabc} \\
\midrule
Multi-image Captioning & The task is to provide multiple images of discrete events and require a title for them. & SSID \citep{abdelhamed2018high} \\

\end{xltabular}

\subsection{Task Abbreviations}
Due to the large number of tasks and models evaluated in the benchmark, we use abbreviations to streamline the manuscript. Table \ref{tab:abbreviation} lists the abbreviations used throughout the paper.

\begin{table*}[ht!]
\centering
\caption{The Abbreviations of terms mentioned in this paper and their corresponding full terms.}
\label{tab:abbreviation}
\resizebox{0.9\textwidth}{!}{%
\begin{tabular}{l|l|l|l}
\toprule
Abbreviation & Full Term & Abbreviation & Full Term \\
\midrule
\multicolumn{4}{c}{Tasks} \\
\midrule
I2IR & Image2Image Retrieval & S2IR & Sketch2Image Retrieval \\
VR & Vehicle Retrieval & FD & Forensic Detection \\
CR & Causality Reasoning & T2IR & Text2Image Retrieval \\
FR & Face Retrieval & ER & Emotion Recognition \\
FC & Functional Correspondence & HR & Handwritten Retrieval \\
VCor & Visual Correspondence & VGR & Visually Grounded Reasoning \\
STD & Spot the Difference & VQA & Visual Quality Assessment \\
MIC & Multiple Image Captioning & PR & Person Re-ID \\
SC & Semantic Correspondence & STS & Spot the Similarity \\
GAR & General Action Recognition & AQA & Action Quality Assessment \\
NIP & Next Image Prediction & MVU & Meme Video Understanding \\
TL & Temporal Localization & MEV & MeVis \\
TC & Textual Cloze & GuAR & GUI App Recognition \\
GNAP & GUI Next Action Prediction & VO & Visual Ordering \\
VCo & Visual Coherence & VidCap & Video Captioning \\
TO & Temporal Ordering & VClz & Visual Cloze \\
MAR & Multiview Action Recognition & HE & Homography Estimation \\
3DOT & 3D Object Tracking & ICSC & Image Captioning with Spatial Context \\
MR & Multiview Reasoning & ITRSC & Image Text Retrieval with Spatial Context \\
IQASC & Icon Question Answering with Spatial Context & 3DE & 3D Depth Estimation \\
RPM & Ravens Progressive Matrices & 3DPE & 3D Pose Estimation \\
3DSR & 3D Scene Reconstruction & JPS & Jigsaw Puzzle Solving \\
3DCR & 3D CAD Recognition & 3DOD & 3D Object Detection \\
ISTE & Image Spatial Transformation Estimation & EVQA & Egocentric Video Question Answering \\
3DIR & 3D Indoor Recognition & 3DQA & 3D Question Answering \\
PT & Point Tracking & SOT & Single Object Tracking \\
\bottomrule
\end{tabular}}
\end{table*}

\subsection{Construction}
\label{appendix:construction}

\textbf{Metadata. } We organize the dataset of each collected task into a metadata format. This structured format helps us easily convert it into multiple-choice questions without losing information. The specific metadata format can be referenced in Table \ref{tab:metadata}.

\begin{table}[h!]
\centering
\begin{tabular}{|p{0.9\textwidth}|}
\rowcolor{black} \color{white}\textbf{Metadata Example} \\
\hline
\vspace{0.02em}

Task Info: \{ \\\quad TaskName: \textit{Name of the task} , \\\quad TaskDescription: \textit{Description of the task} , \\\quad Input Format: \textit{Input data formats, such as text and images} , \\\quad Output Format: \textit{Output data format, such as text or image} ,\\
\}  \\
Samples: $[$ \\
\quad \{ \\
\quad\quad Source: \textit{The data set source of this sample} , \\
\quad\quad Input: \{ \\
\quad\quad\quad Input Image: \textit{The path of the input image, in list format} , \\
\quad\quad\quad Input Context: \textit{The context needed to solve the problem, in text form} , \\
\quad\quad\quad Question: \textit{Input question or instruction} , \\
\quad\quad\quad Visual Component: \textit{Image type, such as depth image, natural image} , \\
\quad\quad\quad \} \\
\quad\quad Output: \textit{The actual textual output of the problem, which may be text (caption task) or image path (retrieval task), etc. } , \\
\quad\}\\
$]$ \\ 
\hline
\end{tabular}
\label{tab:metadata}
\end{table}

\textbf{Unanswerable Set. } We consider five strategies for modifying an answerable instance into its unanswerable counterpart with minimal changes. The five strategies include replacing key words, replacing the answer image, replacing other images, shuffling all images, and using irrelevant question/image sets. For each task, we select the most suitable strategy or combination of strategies to construct the corresponding unanswerable task. The specific construction methods for each task can be referenced in Table \ref{tab:unanswerableall}.

\begin{table}[h]
\centering
\scalebox{0.8}{
\begin{tabular}{lccccc}
\toprule
Task & replace key word & replace answer image & replace other images & shuffle all images & irrelevant question/image set \\
\midrule
CR & Yes & No & No & No & Yes  \\
T2IR & Yes & Yes & No & No & Yes  \\
VCo & No & Yes & No & No & No \\
VO & No & No & No & No & No \\
GAR & No & No & No & Yes & Yes  \\
TL & Yes & Yes & No & No & Yes  \\
TO & No & No & No & Yes & No  \\
VidCap & No & No & No & Yes & Yes  \\
HE & No & No & Yes & Yes & Yes  \\
IQASC & Yes & Yes & Yes & No & Yes  \\
ISTE & No & No & Yes & No & Yes  \\
ITRSC & Yes & Yes & Yes & No & Yes  \\
JPS & No & No & Yes & Yes & Yes  \\
MAR & No & No & Yes & No & Yes  \\
3DE & No & Yes & Yes & Yes & Yes  \\
3DOD & No & No & Yes & Yes & No  \\
3DPE & No & No & Yes & Yes & Yes  \\
3DSR & No & Yes & Yes & Yes & Yes  \\
3DCR & No & Yes & No & No & Yes  \\
3DIR & No & Yes & No & No & Yes  \\
\bottomrule
\end{tabular}}
\caption{Causes of Unanswerable Tasks Considered in Their Construction: Note that a single task often corresponds to multiple causes for being unanswerable.}
\label{tab:unanswerableall}
\end{table}

\section{Experiment Details}
\label{appendix:experiment}

\subsection{Model Details}

Table \ref{tab:lvlm} provides an overview of the LVLMs utilized in this study, detailing their parameter sizes, visual encoders, and LLMs. It is important to mention that the evaluation process was carried out according to the protocol established by OpenCompass. \citep{2023opencompass}

\begin{table*}[!ht]
    \centering
    \caption{Model architecture of $24$ LVLMs evaluated on MMIU.}
    \label{tab:lvlm}
    \scalebox{0.8}
    {
    \begin{tabular}{llll}
        \toprule
        Models & Parameters & Vision Encoder & LLM \\ 
        \midrule
        GPT4o \citep{gpt4o} & - & - & -  \\ 
        Gemini1.5 flash \citep{team2023gemini} & - & - & - \\ 
        Gemini1.0 ProVision \citep{team2023gemini} & - & - & -  \\ 
        Claude3.5-Sonnet \citep{Claude2023} & - & - & -  \\ 
        \midrule
        
        LLaVA-Next-Vicuna-7B \citep{liu2024improved} &7.1B & CLIP ViT-L/14 & Vicuna-v1.5-7B  \\ 
        LLaVA-Next-Interleave \citep{liu2024improved} &7.1B & CLIP ViT-L/14 & Vicuna-v1.5-7B  \\ 
        InternVL2-Pro \citep{chen2024far} & - & - & - \\ 
        InternVL-Chat-V1.5 \citep{chen2024far} & 40B & InternViT-6B & Nous-Hermes-2-Yi-34B \\ 
        DeepSeek-VL-1.3B \citep{lu2024deepseek} & 1.3B & SAM-B \& SigLIP-L & DeekSeek-1.3B  \\  
        DeepSeek-VL-7B \citep{lu2024deepseek} & 7.3B & SAM-B \& SigLIP-L & DeekSeek-7B  \\  
        Monkey-chat \citep{li2024monkey} & 9.8B & CLIP-ViT-BigHuge & Qwen-7B  \\ 
        XComposer2 \citep{dong2024internlm}& 7B & CLIP ViT-L/14 & InternLM2-7B  \\ 
        XComposer2-1.8b \citep{dong2024internlm}& 1.8B & CLIP ViT-L/14 & InternLM2-1.8B  \\ 
        ShareGPT4V \citep{chen2023sharegpt4v}& 7.2B & CLIP ViT-L/14 & Vicuna-v1.5-7B  \\ 
        SharedCaptioner \citep{chen2023sharegpt4v} & 8B & EVA-G & InternLM-7B  \\ 
        LLaVA-v1.5-7B \citep{liu2024llava}& 7.2B & CLIP ViT-L/14 & Vicuna-v1.5-7B  \\ 

        Qwen-base \citep{bai2023qwen} & 9.6B & CLIP ViT-G/16 & Qwen-7B  \\

        Qwen-chat \citep{bai2023qwen} & 9.6B & CLIP ViT-G/16 & Qwen-7B  \\
        Idefics2-8b \citep{laurenccon2024matters} & 8B & SigLIP-L & Mistral-7B  \\
        Idefics-9b-instruct \citep{bai2023qwen} & 9B & CLIP-ViT-H-14 & Llama-7b   \\
        Monkey-chat \citep{li2024monkey} & 9.8B & Vit-BigG & QwenVL-7B   \\
        MiniCPM-Llama3-V-2.5 \citep{hu2024minicpm} & 8.4B & SigLip-L & Llama3-8B   \\
        FlamingoV2 \citep{awadalla2023openflamingo} & 9.6B & CLIP ViT-G/16 & Qwen-7B   \\
        GLM-4V-9B \citep{glm2024chatglm} & 13B & - & GLM-4-9B   \\
        Mantis \citep{jiang2024mantis} & 8B & CLIP ViT-L/14 & Vicuna-v1.5-7B \\
        
        \bottomrule
    \end{tabular}
    }
\end{table*}

\subsection{Model Prompts}
\label{appendix:prompts}
According to MathVista \citep{lu2023mathvista}, our prompt consists of four parts: the question, options, the hint indicating the answer format, and the context of this task (e.g. \textit{Your task is to track the movement of objects in 3D space across multiple frames, select from the following choices.}). For images, we insert them into the text to form a coherent prompt. The complete prompt is as shown in Table \ref{tab:prompts}.

\begin{table}[h!]
\centering
\begin{tabular}{|p{0.9\textwidth}|}
\rowcolor{black} \color{white}\textbf{Model Prompts} \\
\hline
\vspace{0.02em}

Context: \{CONTEXT\}  \\
Question: \{QUESTION\} \\ 
Choices: \\
(A) \{OPTION\_A\} \\
(B) \{OPTION\_B\} \\
(C) \{OPTION\_C\} \\
(D) \{OPTION\_D\} \\
Hint: Please answer the option directly like A, B, C, D... \\
\hline
\end{tabular}
\label{tab:prompts}
\end{table}

\subsection{Multitask Analysis}

\begin{figure*}
    \centering
    \includegraphics[width=0.9\linewidth]{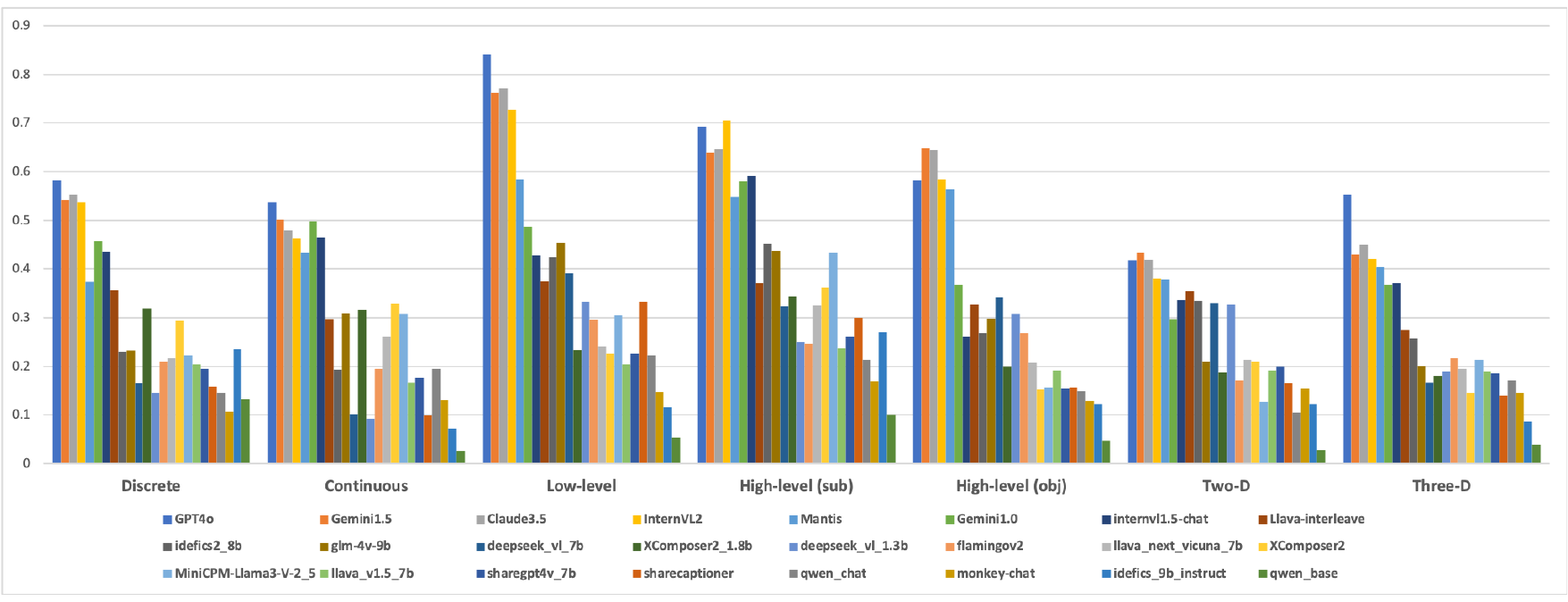}
    \vspace{-1.0in}
    \caption{The average performance comparison of 24 LVLMs on seven specific image relationships.}
    \label{fig:meta_all}
\end{figure*}

\textbf{Differences in Model Capabilities Across Various Image Relationships.}
As shown in Figure \ref{fig:meta_all}, we visualize the average performance of all models across 7 specific image relationships. Detailed analysis can be found in Sec .\ref{sec:multitask} of the main text.

\subsection{Error Analysis}
\begin{figure*}
    \centering
    \includegraphics[width=0.85\linewidth]{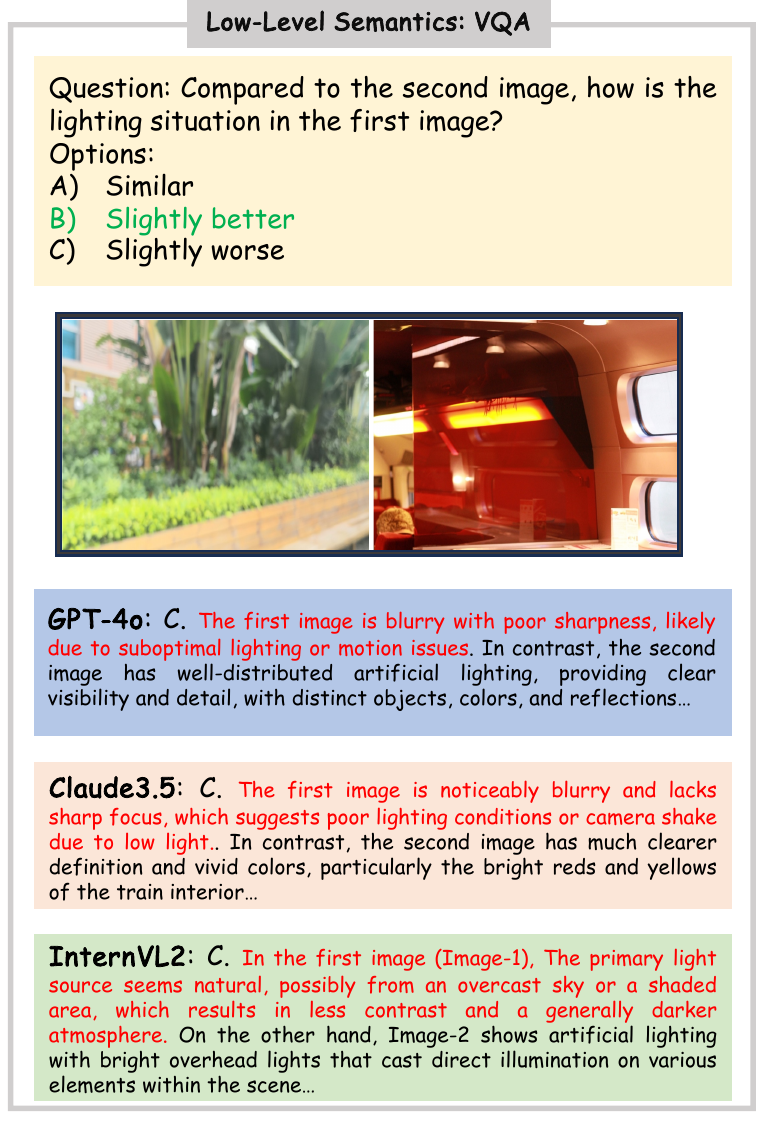}
    \caption{Error cases in low-level semantic image relations (visual quality assessment). Red highlights the wrong answer.}
    \label{fig:error_lowlevel}
\end{figure*}

\begin{figure*}
    \centering
    \includegraphics[width=0.9\linewidth]{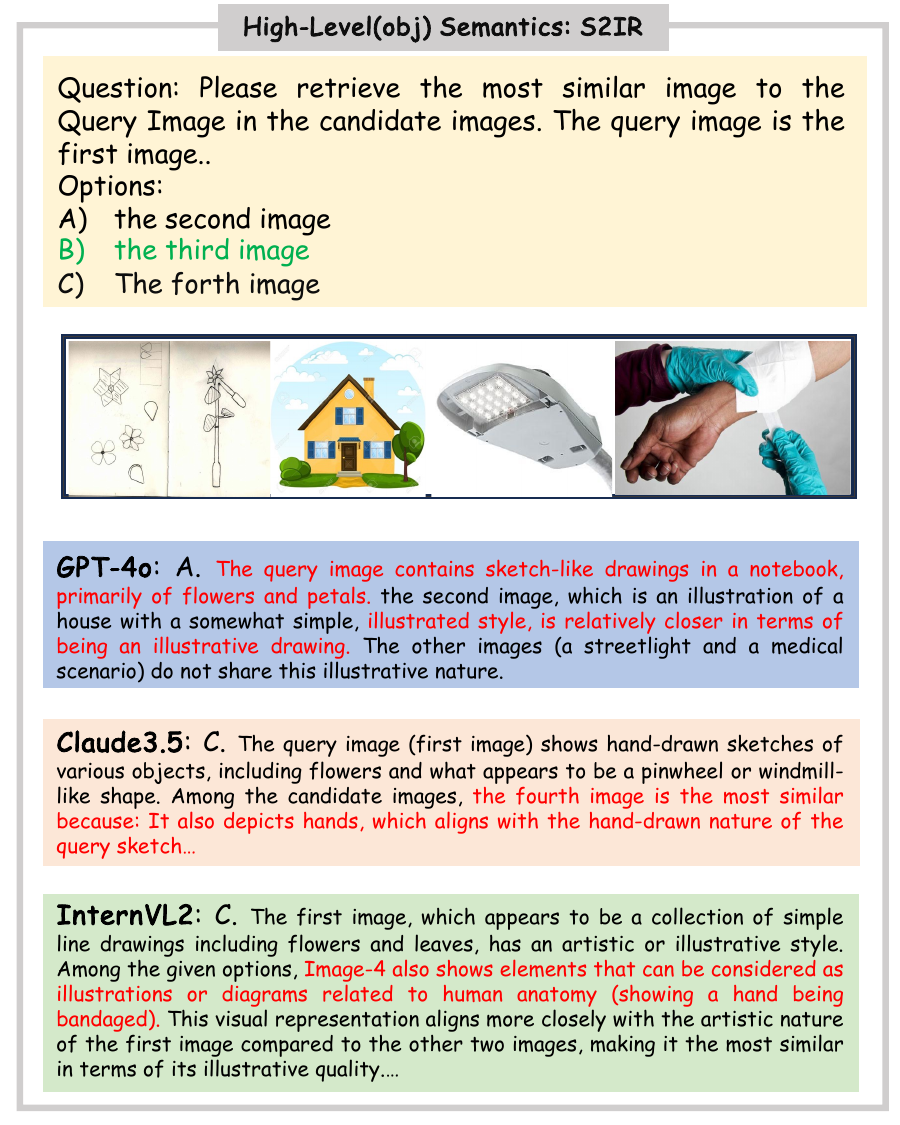}
    \caption{An error case in high-level(obj) semantic image relations (sketch2image retrieval). Red highlights the wrong answer.}
    \label{fig:error_highlevelobj}
\end{figure*}

\begin{figure*}
    \centering
    \includegraphics[width=0.9\linewidth]{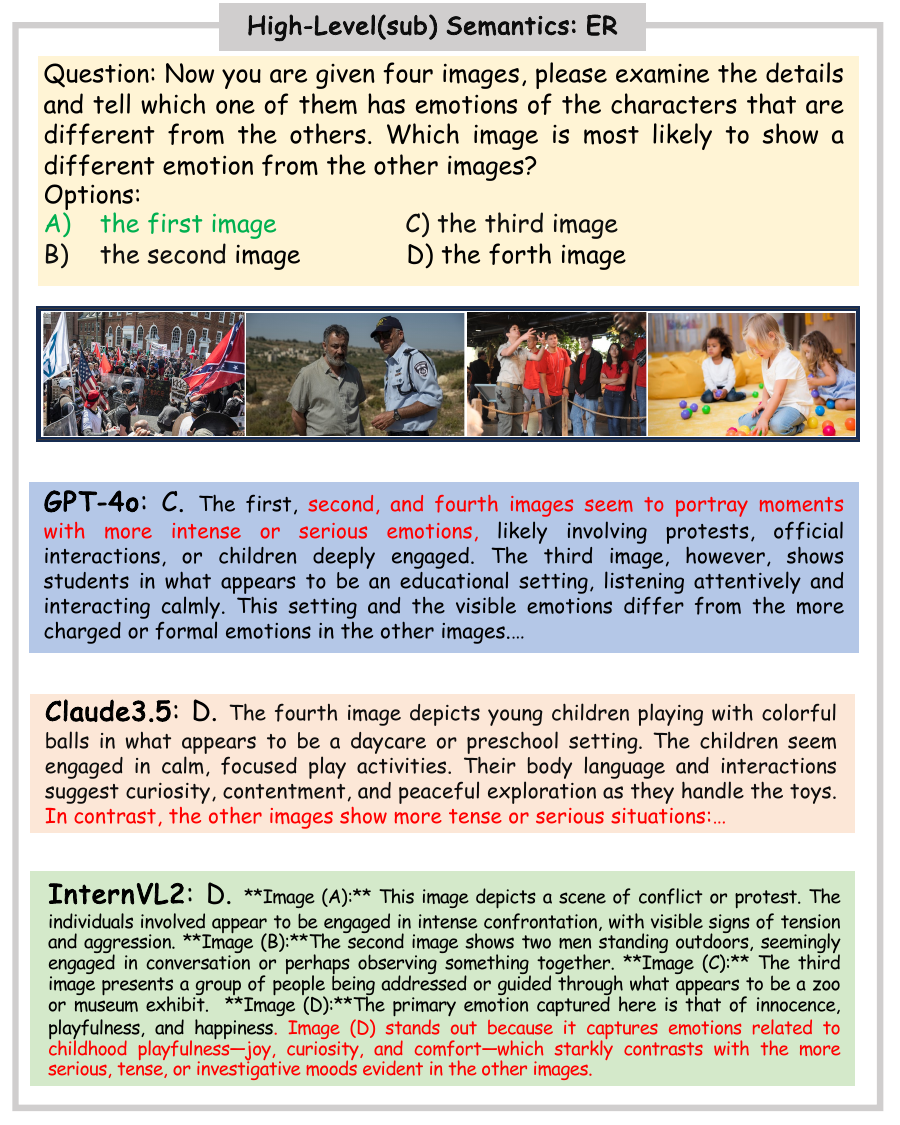}
    \caption{An error case in high-level(sub) semantic image relations (emotion recognition). Red highlights the wrong answer.}
    \label{fig:error_highlevelsub}
\end{figure*}

\begin{figure*}
    \centering
    \includegraphics[width=0.9\linewidth]{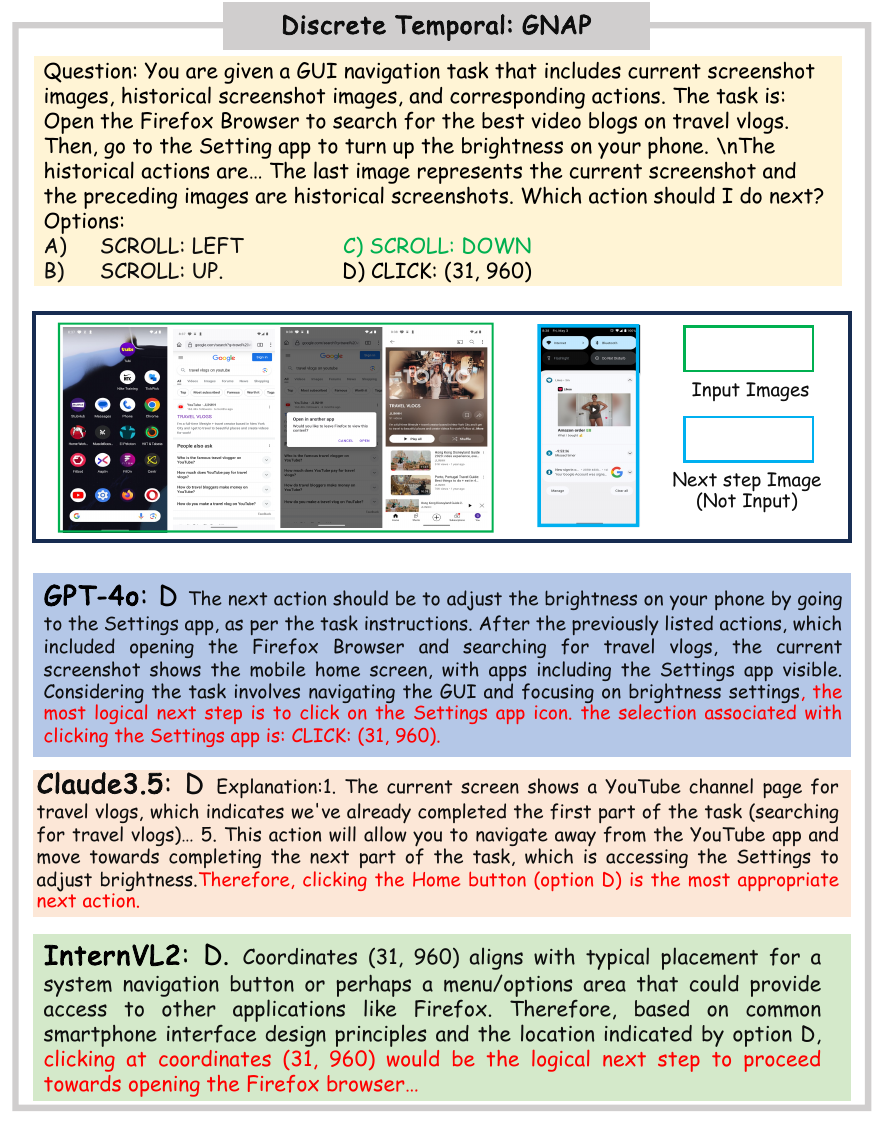}
    \caption{An error case in discrete temporal image relations (GUI APP action prediction). Red highlights the wrong answer.}
    \label{fig:error_discrete}
\end{figure*}

\begin{figure*}
    \centering
    \includegraphics[width=0.9\linewidth]{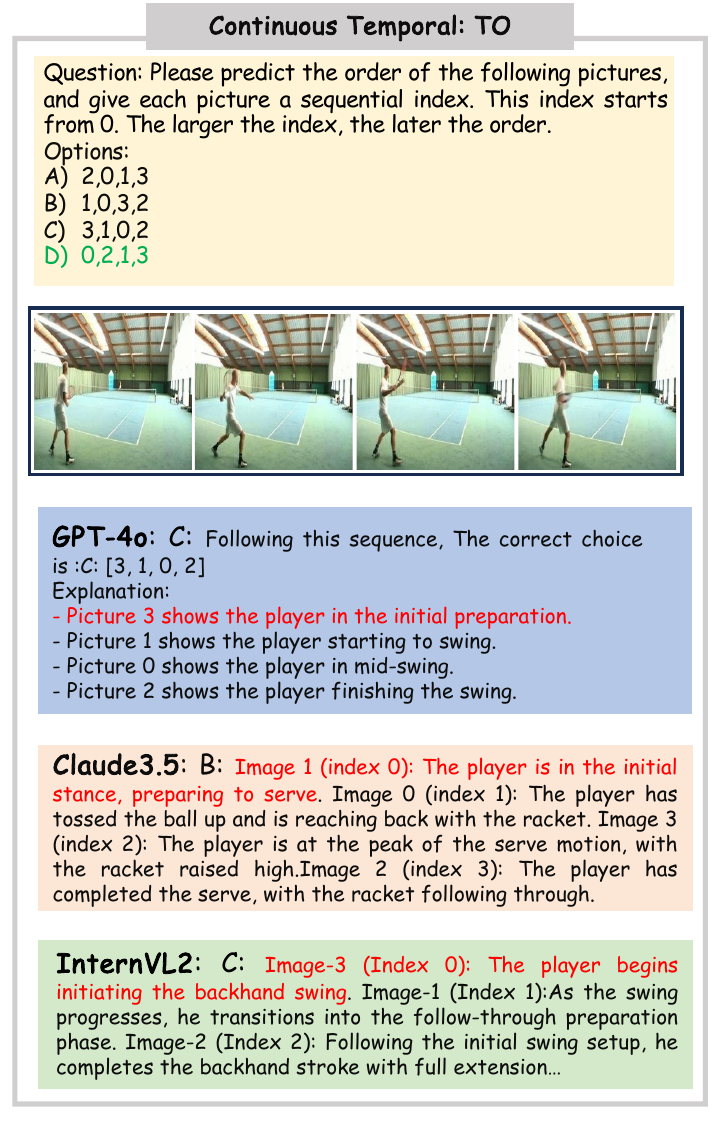}
    \caption{An error case in continuous temporal image relations (temporal ordering). Red highlights the wrong answer.}
    \label{fig:error_continuous}
\end{figure*}

\begin{figure*}
    \centering
    \includegraphics[width=0.9\linewidth]{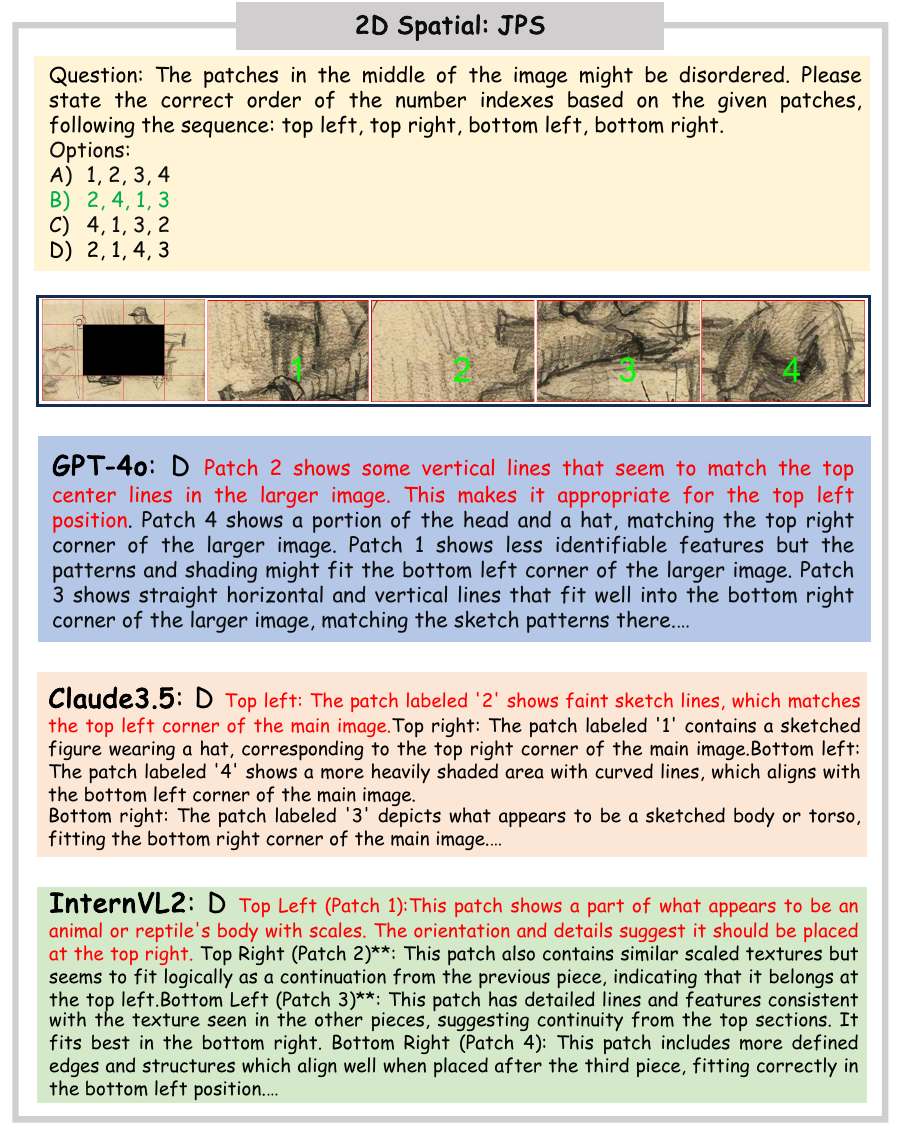}
    \caption{An error case in 2D spatial image relations (jigsaw puzzle solving). Red highlights the wrong answer.}
    \label{fig:error_2d}
\end{figure*}

\begin{figure*}
    \centering
    \includegraphics[width=0.9\linewidth]{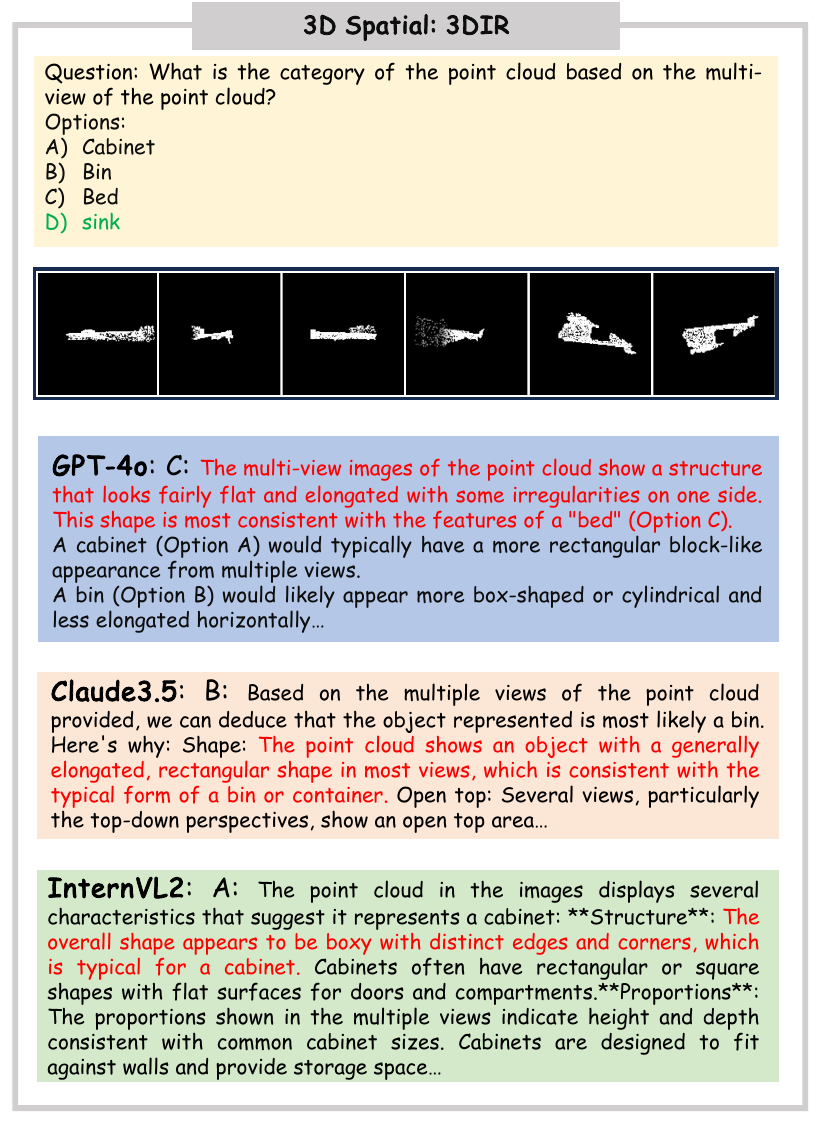}
    \caption{An error case in 3D spatial image relations (3D indoor recognition). Red highlights the wrong answer.}
    \label{fig:error_3d}
\end{figure*}

To more vividly demonstrate the performance of MMIU and the models on different tasks, we analyze error cases for the three best-performing models: GPT-4o, Claude3.5-Sonnet, and InternVL2-Pro. For each type of image relationship, we visualize one error case. Since the model responses or questions might be lengthy, we mostly extract the important parts. The images are shown in Figure \ref{fig:error_lowlevel}, Figure \ref{fig:error_highlevelobj}, Figure \ref{fig:error_highlevelsub}, Figure \ref{fig:error_discrete}, Figure \ref{fig:error_continuous}, Figure \ref{fig:error_2d}, and Figure \ref{fig:error_3d}.

\subsection{Ablation Study}

\textbf{Impact of Setting Unanswerable Questions on Model Performance.}
We test the performance of most representative LVLMs on 19 tasks with both answerable and unanswerable types. We summarize all the results in Table \ref{tab:unanswerableall}.

\begin{table*}[t!]
\centering
\caption{Quantitative results for LVLMs across 21 tasks with unanswerable and answerable questions are summarized. Accuracy is the metric, and the Overall score is computed across all tasks.}
\label{tab:unanswerableall}
\resizebox{0.99\textwidth}{!}
{
\begin{tabular}{l|lllllllllllllllllllllllllll}
\toprule
Model & Overall & T2IR & VCo & VO & GAR & TL & TO & VidCap & HE & IQASC & ISTE & ITRSC & MAR & JPS & 3DE & 3DOD & 3DPE & 3DSR & 3DCR & 3DIR & CR \\
\midrule
\multicolumn{21}{c}{\multirow{2}{*}{\textbf{Answerable}}} \\
\multicolumn{21}{c}{} \\
\midrule
GPT4o & 54.2 & 70.5 & 97.5 & 21.5 & 91.5 & 50.8 & 28.0 & 92.5 & 29.5 & 88.0 & 35.0 & 17.5 & 81.9 & 23.5 & 24.0 & 40.5 & 85.0 & 22.0 & 69.0 & 49.0 & 67.8 \\
\midrule
Claude3.5 & 49.5 & 65.5 & 67.5 & 38.5 & 80.5 & 43.5 & 23.0 & 91.0 & 23.0 & 78.5 & 32.0 & 4.5 & 64.8 & 31.5 & 23.5 & 41.0 & 99.5 & 21.5 & 53.5 & 36.5 & 70.2 \\
\midrule
Mantis & 38.9 & 46.5 & 14.0 & 20.0 & 81.0 & 23.8 & 27.0 & 85.0 & 23.0 & 66.0 & 23.5 & 13.0 & 71.4 & 27.5 & 23.5 & 24.0 & 22.5 & 25.0 & 59.0 & 40.5 & 61.5 \\
\midrule
internvl1.5-chat & 37.5 & 26.5 & 31.0 & 24.5 & 85.0 & 29.0 & 18.5 & 89.0 & 26.0 & 40.0 & 25.5 & 15.5 & 59.3 & 19.5 & 22.5 & 23.5 & 33.5 & 28.0 & 50.5 & 39.5 & 63.7 \\
\midrule
idefics2-8b & 24.0 & 12.5 & 21.5 & 22.5 & 24.5 & 22.3 & 18.0 & 19.5 & 21.5 & 31.0 & 25.5 & 13.5 & 15.1 & 27.5 & 26.0 & 21.5 & 21.5 & 23.0 & 44.5 & 40.5 & 28.0 \\
\midrule
glm-4v-9b & 23.3 & 25.0 & 11.5 & 14.5 & 59.5 & 25.9 & 10.0 & 76.0 & 11.5 & 35.5 & 16.0 & 6.5 & 25.1 & 9.0 & 14.0 & 14.5 & 0.5 & 5.5 & 48.5 & 23.5 & 32.8 \\
\midrule
deepseek-vl-7b & 19.3 & 25.5 & 30.5 & 21.5 & 6.0 & 31.1 & 15.5 & 2.0 & 23.0 & 45.5 & 24.5 & 0.0 & 2.0 & 20.5 & 24.5 & 24.5 & 7.5 & 0.5 & 40.5 & 38.5 & 2.2 \\
\midrule
XComposer2-1.8b & 19.5 & 10.5 & 28.5 & 17.0 & 59.5 & 24.4 & 13.0 & 68.5 & 0.5 & 29.5 & 6.0 & 7.5 & 33.2 & 7.0 & 0.0 & 15.5 & 28.0 & 2.0 & 11.5 & 3.0 & 24.5 \\
\midrule
deepseek-vl-1.3b & 20.1 & 23.0 & 33.0 & 20.0 & 1.0 & 26.4 & 20.0 & 1.0 & 25.0 & 44.5 & 24.0 & 1.0 & 0.0 & 31.0 & 26.0 & 31.0 & 19.5 & 0.0 & 45.5 & 29.0 & 1.2 \\
\midrule
flamingov2 & 19.0 & 20.0 & 14.5 & 13.5 & 13.5 & 18.7 & 24.5 & 22.5 & 27.5 & 4.0 & 23.0 & 7.0 & 22.1 & 1.5 & 26.5 & 22.0 & 17.0 & 28.5 & 25.0 & 23.5 & 25.5 \\
\midrule
llava-next-vicuna-7b & 18.1 & 13.0 & 9.5 & 8.5 & 51.0 & 17.1 & 6.5 & 66.0 & 5.0 & 28.5 & 8.5 & 5.0 & 22.6 & 6.5 & 4.0 & 4.0 & 8.0 & 9.5 & 42.0 & 25.0 & 22.2 \\
\midrule
XComposer2 & 17.8 & 9.5 & 2.5 & 19.0 & 59.5 & 15.5 & 12.0 & 66.0 & 8.0 & 15.5 & 0.0 & 0.0 & 20.6 & 16.5 & 0.0 & 7.0 & 4.5 & 0.0 & 42.0 & 33.0 & 24.0 \\
\midrule
MiniCPM-Llama3-V-2-5 & 20.6 & 10.0 & 20.5 & 12.0 & 70.5 & 17.6 & 5.0 & 83.5 & 0.0 & 25.0 & 0.0 & 1.5 & 34.2 & 6.0 & 8.5 & 5.5 & 20.0 & 4.5 & 32.5 & 15.0 & 41.1 \\
\midrule
llava-v1.5-7b & 14.8 & 19.0 & 21.5 & 4.0 & 23.5 & 6.7 & 7.0 & 28.0 & 7.5 & 26.5 & 5.0 & 4.5 & 25.6 & 8.5 & 8.0 & 4.0 & 6.0 & 14.5 & 34.5 & 28.5 & 14.1 \\
\midrule
sharegpt4v-7b & 14.3 & 13.5 & 14.0 & 7.5 & 26.0 & 10.9 & 6.0 & 25.0 & 7.0 & 29.0 & 5.0 & 1.5 & 28.1 & 9.5 & 3.0 & 7.0 & 2.0 & 8.0 & 36.5 & 31.0 & 16.4 \\
\midrule
sharecaptioner & 11.3 & 12.5 & 14.5 & 11.0 & 14.5 & 18.1 & 5.5 & 21.5 & 7.0 & 25.5 & 5.5 & 2.0 & 16.1 & 9.0 & 2.5 & 1.5 & 5.5 & 8.0 & 16.5 & 9.0 & 20.7 \\
\midrule
qwen-chat & 15.5 & 3.0 & 6.0 & 6.0 & 52.5 & 3.6 & 5.5 & 47.0 & 9.0 & 13.5 & 15.5 & 3.5 & 40.2 & 16.5 & 16.5 & 22.5 & 13.0 & 14.5 & 1.5 & 0.5 & 20.5 \\
\midrule
monkey-chat & 11.8 & 12.0 & 13.0 & 7.5 & 16.5 & 18.1 & 6.5 & 19.5 & 7.0 & 27.5 & 5.5 & 3.0 & 10.6 & 9.0 & 5.5 & 8.0 & 5.5 & 7.5 & 36.0 & 8.5 & 8.4 \\
\midrule
idefics-9b-instruct & 6.1 & 3.5 & 0.0 & 5.5 & 1.5 & 0.5 & 3.0 & 10.0 & 3.0 & 9.0 & 0.0 & 0.0 & 6.5 & 1.0 & 15.5 & 10.5 & 36.5 & 5.5 & 0.0 & 0.0 & 10.8 \\
\midrule
qwen-base & 3.7 & 1.0 & 5.5 & 6.5 & 4.5 & 0.0 & 0.0 & 7.5 & 2.0 & 8.5 & 0.0 & 0.0 & 0.5 & 0.0 & 0.5 & 7.0 & 21.5 & 0.0 & 0.0 & 0.0 & 9.2 \\

\midrule
\multicolumn{21}{c}{\multirow{2}{*}{\textbf{Unanswerable}}} \\
\multicolumn{21}{c}{} \\
\midrule
GPT4o & 43.4 & 32.5 & 72.5 & 5.0 & 65.0 & 51.3 & 7.5 & 75.0 & 92.5 & 32.5 & 32.5 & 57.5 & 30.0 & 72.5 & 5.0 & 67.5 & 95.0 & 5.0 & 25.0 & 7.5 & 37.5 \\
\midrule
Claude3.5 & 36.2 & 45.0 & 80.0 & 0.0 & 62.5 & 38.5 & 0.0 & 70.0 & 65.0 & 35.0 & 30.0 & 62.2 & 62.5 & 5.0 & 2.5 & 32.5 & 42.5 & 5.0 & 20.0 & 7.5 & 58.8 \\
\midrule
Mantis & 48.3 & 17.5 & 7.5 & 22.5 & 62.5 & 61.5 & 0.0 & 80.0 & 75.0 & 37.5 & 50.0 & 2.5 & 35.0 & 2.5 & 87.5 & 67.5 & 57.5 & 57.5 & 82.5 & 87.5 & 72.5 \\
\midrule
internvl1.5-chat & 37.7 & 15.0 & 75.0 & 0.0 & 55.0 & 35.9 & 2.5 & 70.0 & 20.0 & 12.5 & 10.0 & 7.5 & 37.5 & 0.0 & 82.5 & 80.0 & 65.0 & 85.0 & 20.0 & 42.5 & 37.5 \\
\midrule
idefics2-8b & 26.4 & 5.0 & 90.0 & 0.0 & 30.0 & 20.5 & 2.5 & 32.5 & 0.0 & 45.0 & 2.5 & 27.5 & 25.0 & 0.0 & 100.0 & 27.5 & 0.0 & 32.5 & 20.0 & 32.5 & 35.0 \\
\midrule
glm-4v-9b & 9.1 & 5.0 & 10.0 & 0.0 & 12.5 & 28.2 & 0.0 & 30.0 & 0.0 & 15.0 & 0.0 & 7.5 & 0.0 & 0.0 & 0.0 & 12.5 & 0.0 & 0.0 & 10.0 & 20.0 & 31.2 \\
\midrule
deepseek-vl-7b & 20.0 & 0.0 & 2.5 & 2.5 & 37.5 & 0.0 & 0.0 & 22.5 & 2.5 & 25.0 & 0.0 & 27.5 & 27.5 & 0.0 & 67.5 & 22.5 & 25.0 & 5.0 & 42.5 & 77.5 & 12.5 \\
\midrule
XComposer2-1.8b & 39.9 & 17.5 & 0.0 & 2.5 & 25.0 & 2.6 & 32.5 & 47.5 & 52.5 & 47.5 & 47.5 & 27.5 & 27.5 & 2.5 & 92.5 & 62.5 & 55.0 & 27.5 & 95.0 & 95.0 & 38.8 \\
\midrule
deepseek-vl-1.3b & 21.6 & 2.5 & 27.5 & 7.5 & 22.5 & 0.0 & 10.0 & 10.0 & 0.0 & 17.5 & 15.0 & 15.0 & 30.0 & 0.0 & 37.5 & 35.0 & 25.0 & 7.5 & 77.5 & 75.0 & 16.2 \\
\midrule
flamingov2 & 23.2 & 5.0 & 17.5 & 2.5 & 72.5 & 0.0 & 0.0 & 62.5 & 35.0 & 42.5 & 0.0 & 42.5 & 20.0 & 0.0 & 27.5 & 0.0 & 0.0 & 12.5 & 40.0 & 32.5 & 51.2 \\
\midrule
llava-next-vicuna-7b & 6.6 & 0.0 & 0.0 & 0.0 & 27.5 & 0.0 & 0.0 & 30.0 & 0.0 & 0.0 & 0.0 & 0.0 & 7.5 & 0.0 & 0.0 & 0.0 & 0.0 & 0.0 & 10.0 & 47.5 & 8.8 \\
\midrule
XComposer2 & 38.0 & 35.0 & 50.0 & 12.5 & 52.5 & 35.9 & 0.0 & 70.0 & 30.0 & 65.0 & 5.0 & 30.0 & 80.0 & 0.0 & 47.5 & 22.5 & 25.0 & 42.5 & 47.5 & 50.0 & 60.0 \\
\midrule
MiniCPM-Llama3-V-2-5 & 13.1 & 5.0 & 15.0 & 0.0 & 42.5 & 23.1 & 0.0 & 47.5 & 15.0 & 7.5 & 10.0 & 0.0 & 5.0 & 0.0 & 0.0 & 0.0 & 0.0 & 0.0 & 47.5 & 20.0 & 23.8 \\
\midrule
llava-v1.5-7b & 19.1 & 0.0 & 0.0 & 0.0 & 50.0 & 0.0 & 0.0 & 37.5 & 0.0 & 0.0 & 20.0 & 12.5 & 15.0 & 0.0 & 30.0 & 12.5 & 2.5 & 10.0 & 60.0 & 85.0 & 46.2 \\
\midrule
sharegpt4v-7b & 19.1 & 0.0 & 0.0 & 0.0 & 40.0 & 0.0 & 0.0 & 15.0 & 7.5 & 12.5 & 17.5 & 17.5 & 10.0 & 0.0 & 32.5 & 5.0 & 0.0 & 15.0 & 77.5 & 87.5 & 43.8 \\
\midrule
sharecaptioner & 10.8 & 0.0 & 0.0 & 0.0 & 2.5 & 0.0 & 0.0 & 10.0 & 0.0 & 2.5 & 0.0 & 0.0 & 2.5 & 0.0 & 0.0 & 12.5 & 0.0 & 0.0 & 70.0 & 80.0 & 36.2 \\
\midrule
qwen-chat & 22.0 & 12.5 & 10.0 & 0.0 & 60.0 & 25.6 & 25.0 & 60.0 & 7.5 & 10.0 & 17.5 & 35.0 & 17.5 & 20.0 & 32.5 & 20.0 & 17.5 & 22.5 & 12.5 & 2.5 & 31.2 \\
\midrule
monkey-chat & 12.2 & 0.0 & 0.0 & 0.0 & 2.5 & 0.0 & 0.0 & 20.0 & 7.5 & 12.5 & 0.0 & 2.5 & 15.0 & 0.0 & 25.0 & 5.0 & 0.0 & 10.0 & 42.5 & 75.0 & 26.2 \\
\midrule
idefics-9b-instruct & 56.1 & 77.5 & 52.5 & 85.0 & 80.0 & 86.8 & 60.0 & 87.5 & 30.0 & 57.5 & 45.0 & 97.5 & 60.0 & 27.5 & 42.5 & 7.5 & 27.5 & 12.5 & 67.5 & 60.0 & 57.5 \\
\midrule
qwen-base & 44.4 & 72.5 & 67.5 & 27.5 & 65.0 & 79.5 & 60.0 & 45.0 & 47.5 & 50.0 & 30.0 & 65.0 & 95.0 & 62.5 & 17.5 & 30.0 & 5.0 & 0.0 & 2.5 & 5.0 & 61.3 \\
  
  \bottomrule
\end{tabular}
}
\end{table*}

\textbf{Impact of Different Testing Methods on Model Performance.}
We test the effectiveness of single-image input models in completing multi-image tasks using concatenated visual tokens or concatenated images, and we record the final results in Table \ref{tab:tokenall}.

\begin{table*}[t!]
\centering
\caption{Quantitative results for single-image LVLMs across 52 mtasks with token-concat or image-concat are summarized. Accuracy is the metric, and the Overall score is computed across all tasks.}
\label{tab:tokenall}
\resizebox{0.99\textwidth}{!}
{
\begin{tabular}{l|lllllllllllllllllllllllllll}
\toprule
Model & Overall & CR & ER & FD & FC & SC & VCor & VQA & VGR & FR & HR & I2IR & MIC & PR & S2IR & STD & STS & T2IR & VR & AQA & GAR & MVU & MEV & NIP & TL & TO & VidCap \\
  &   & GuAR & GNAP & TC & VClz & VCo & VO & EVQA & HE & IQASC & ICSC & ISTE & ITRSC & MAR & MR & JPS & 3DE & 3DOD & 3DOT & 3DPE & 3DSR & 3DQA & PT & RPM & SOT & 3DCR & 3DIR \\
\midrule
\multicolumn{28}{c}{\multirow{2}{*}{\textbf{image-concat}}} \\
\multicolumn{28}{c}{} \\
\midrule
glm-4v-9b & 27.0 & 32.8 & 16.0 & 31.8 & 8.7 & 9.0 & 4.7 & 59.0 & 55.8 & 31.0 & 7.5 & 19.5 & 82.0 & 23.5 & 24.5 & 81.0 & 67.0 & 25.0 & 30.0 & 7.0 & 59.5 & 53.5 & 10.5 & 5.0 & 25.9 & 10.0 & 76.0 \\
  &   & 55.5 & 19.0 & 34.0 & 5.0 & 11.5 & 14.5 & 26.0 & 11.5 & 35.5 & 41.5 & 16.0 & 6.5 & 25.1 & 29.3 & 9.0 & 14.0 & 14.5 & 7.0 & 0.5 & 5.5 & 27.0 & 35.0 & 7.5 & 26.0 & 48.5 & 23.5 \\
\midrule
llava-next-vicuna-7b & 22.2 & 22.2 & 9.2 & 11.0 & 9.1 & 7.7 & 10.5 & 37.0 & 23.2 & 7.0 & 16.5 & 8.0 & 66.0 & 5.0 & 23.5 & 88.0 & 42.5 & 13.0 & 14.5 & 5.5 & 51.0 & 42.5 & 9.5 & 10.0 & 17.1 & 6.5 & 66.0 \\
  &   & 50.5 & 14.5 & 38.0 & 9.0 & 9.5 & 8.5 & 31.0 & 5.0 & 28.5 & 27.0 & 8.5 & 5.0 & 22.6 & 29.3 & 6.5 & 4.0 & 4.0 & 6.0 & 8.0 & 9.5 & 32.5 & 72.0 & 1.0 & 38.0 & 42.0 & 25.0 \\
\midrule
llava-v1.5-7b & 19.2 & 14.1 & 4.2 & 13.7 & 5.8 & 1.9 & 6.9 & 27.3 & 35.0 & 6.5 & 12.5 & 12.5 & 53.0 & 10.0 & 25.5 & 66.5 & 43.0 & 19.0 & 3.5 & 2.5 & 23.5 & 36.5 & 12.0 & 16.5 & 6.7 & 7.0 & 28.0 \\
  &   & 24.5 & 17.5 & 40.0 & 15.0 & 21.5 & 4.0 & 26.0 & 7.5 & 26.5 & 17.5 & 5.0 & 4.5 & 25.6 & 27.1 & 8.5 & 8.0 & 4.0 & 6.0 & 6.0 & 14.5 & 29.5 & 66.0 & 2.0 & 35.0 & 34.5 & 28.5 \\
\midrule
sharegpt4v-7b & 18.5 & 16.4 & 5.0 & 10.8 & 6.2 & 9.0 & 2.7 & 34.2 & 28.5 & 4.5 & 10.5 & 3.5 & 57.0 & 4.0 & 12.5 & 55.5 & 44.5 & 13.5 & 5.0 & 5.0 & 26.0 & 38.0 & 14.0 & 15.5 & 10.9 & 6.0 & 25.0 \\
  &   & 26.5 & 19.0 & 42.0 & 7.5 & 14.0 & 7.5 & 31.5 & 7.0 & 29.0 & 18.0 & 5.0 & 1.5 & 28.1 & 23.3 & 9.5 & 3.0 & 7.0 & 6.0 & 2.0 & 8.0 & 27.5 & 65.5 & 0.0 & 44.0 & 36.5 & 31.0 \\
\midrule
sharecaptioner & 16.1 & 20.7 & 22.2 & 27.2 & 10.2 & 9.1 & 21.0 & 39.5 & 37.0 & 7.0 & 5.0 & 6.0 & 47.0 & 5.0 & 17.0 & 25.0 & 35.5 & 12.5 & 13.0 & 5.5 & 14.5 & 4.5 & 3.0 & 6.0 & 18.1 & 5.5 & 21.5 \\
  &   & 17.0 & 22.5 & 18.5 & 12.0 & 14.5 & 11.0 & 23.5 & 7.0 & 25.5 & 22.0 & 5.5 & 2.0 & 16.1 & 43.6 & 9.0 & 2.5 & 1.5 & 1.5 & 5.5 & 8.0 & 26.5 & 47.0 & 2.0 & 28.0 & 16.5 & 9.0 \\
\midrule
monkey-chat & 13.7 & 8.4 & 8.0 & 5.9 & 9.2 & 6.7 & 8.1 & 23.5 & 25.3 & 4.5 & 6.0 & 1.5 & 34.5 & 2.0 & 9.0 & 40.5 & 40.5 & 12.0 & 2.5 & 6.5 & 16.5 & 14.5 & 10.0 & 12.5 & 18.1 & 6.5 & 19.5 \\
  &   & 10.0 & 8.5 & 17.0 & 8.0 & 13.0 & 7.5 & 15.5 & 7.0 & 27.5 & 17.0 & 5.5 & 3.0 & 10.6 & 22.6 & 9.0 & 5.5 & 8.0 & 6.0 & 5.5 & 7.5 & 34.5 & 51.0 & 1.5 & 17.0 & 36.0 & 8.5 \\
\midrule
\multicolumn{28}{c}{\multirow{2}{*}{\textbf{token-concat}}} \\
\multicolumn{28}{c}{} \\
\midrule
glm-4v-9b & 26.7 & 61.2 & 9.8 & 14.1 & 9.1 & 12.2 & 14.4 & 27.5 & 54.0 & 13.0 & 18.0 & 9.0 & 79.5 & 12.5 & 19.5 & 64.5 & 37.5 & 12.5 & 15.0 & 7.5 & 81.5 & 63.0 & 11.5 & 9.0 & 14.0 & 6.5 & 91.5 \\
  &   & 55.0 & 22.0 & 41.0 & 5.0 & 4.5 & 3.5 & 53.0 & 13.0 & 39.5 & 41.0 & 16.0 & 6.0 & 76.9 & 29.3 & 8.0 & 4.5 & 11.0 & 5.5 & 1.0 & 7.0 & 34.5 & 42.0 & 3.5 & 25.5 & 49.5 & 21.5 \\
\midrule
llava-next-vicuna-7b & 5.9 & 0.0 & 0.5 & 0.0 & 4.7 & 0.0 & 0.9 & 1.2 & 11.0 & 0.5 & 0.0 & 0.0 & 0.5 & 0.0 & 2.5 & 62.5 & 45.0 & 0.0 & 0.0 & 0.0 & 0.0 & 0.5 & 0.0 & 18.5 & 14.0 & 0.0 & 0.0 \\
  &   & 2.0 & 2.5 & 1.0 & 0.0 & 2.5 & 1.0 & 0.0 & 0.5 & 10.5 & 34.5 & 0.0 & 0.0 & 0.0 & 0.0 & 0.0 & 0.0 & 0.0 & 0.0 & 0.0 & 0.0 & 0.0 & 77.0 & 0.0 & 12.5 & 0.0 & 0.0 \\
\midrule
llava-v1.5-7b & 13.5 & 0.0 & 4.5 & 10.4 & 10.2 & 8.8 & 7.5 & 30.2 & 20.0 & 11.0 & 19.0 & 10.0 & 58.5 & 18.0 & 30.5 & 35.5 & 45.5 & 25.5 & 14.5 & 0.0 & 0.0 & 0.0 & 0.0 & 15.5 & 2.1 & 5.5 & 0.0 \\
  &   & 2.0 & 8.0 & 4.0 & 2.5 & 27.0 & 8.5 & 0.0 & 7.5 & 7.0 & 26.5 & 6.5 & 0.5 & 0.0 & 29.3 & 8.0 & 17.5 & 5.5 & 0.0 & 11.0 & 2.0 & 0.5 & 69.5 & 0.0 & 36.0 & 26.0 & 12.0 \\
\midrule
sharegpt4v-7b & 14.0 & 0.2 & 4.5 & 15.9 & 9.2 & 9.2 & 5.1 & 26.8 & 27.3 & 13.5 & 19.5 & 21.5 & 58.0 & 15.0 & 22.5 & 30.5 & 45.0 & 25.5 & 22.5 & 0.0 & 0.0 & 0.0 & 0.0 & 18.0 & 5.7 & 7.5 & 0.0 \\
  &   & 2.0 & 3.0 & 4.0 & 1.0 & 24.0 & 5.5 & 0.0 & 7.5 & 6.5 & 28.5 & 7.0 & 0.0 & 0.0 & 29.3 & 6.0 & 17.0 & 5.0 & 0.0 & 0.0 & 1.5 & 2.5 & 75.0 & 0.0 & 40.5 & 34.5 & 26.5 \\
\midrule
sharecaptioner & 9.4 & 0.3 & 8.0 & 13.5 & 16.4 & 11.1 & 21.5 & 23.8 & 33.0 & 0.0 & 0.0 & 1.0 & 35.5 & 0.5 & 1.5 & 4.0 & 3.5 & 10.5 & 1.0 & 0.0 & 0.5 & 2.5 & 0.0 & 5.0 & 0.0 & 8.5 & 0.0 \\
  &   & 6.0 & 12.0 & 2.5 & 9.0 & 18.0 & 10.5 & 0.0 & 7.0 & 19.5 & 30.0 & 6.0 & 1.5 & 0.0 & 22.6 & 7.5 & 2.5 & 4.0 & 0.0 & 6.5 & 0.0 & 0.0 & 62.0 & 0.0 & 14.0 & 22.5 & 23.5 \\
\midrule
monkey-chat & 11.1 & 0.0 & 2.5 & 9.4 & 10.6 & 12.4 & 8.1 & 19.8 & 34.2 & 5.5 & 7.5 & 2.5 & 46.5 & 4.5 & 13.5 & 28.0 & 42.0 & 17.0 & 10.5 & 0.0 & 0.0 & 0.0 & 10.5 & 8.5 & 5.2 & 9.0 & 0.0 \\
  &   & 7.5 & 7.0 & 9.0 & 8.5 & 7.5 & 7.5 & 0.0 & 6.5 & 30.0 & 25.5 & 5.5 & 0.5 & 0.5 & 41.4 & 5.5 & 5.5 & 5.0 & 7.0 & 5.5 & 8.5 & 0.0 & 42.0 & 0.0 & 10.5 & 16.5 & 4.5 \\
  
  \bottomrule
\end{tabular}
}
\end{table*}

\section{Task Map}
\label{appendix:taskmap}
A task map determines the similarity between tasks based on their inherent characteristics. By combining the task map with model performance, we aim to analyze which types of tasks current models perform well or poorly on. This approach avoids the bias introduced by using meta-task analysis alone, providing a more comprehensive conclusion through complementary methods. Thanks to the extensive number of tasks in MMIU, we can construct a comprehensive multi-image task map.

\subsection{Construction}

Inspired by the methodology outlined in TaskVec \citep{ilharco2022taskvector}, and benefiting from the extensive and diverse set of tasks in MMIU, we aim to analyze tasks and model performance across different tasks using a task map. Specifically:
1) We extract task vectors similarly to the approach in TaskVec. Using QwenVL-chat as a probing network, we fine-tune it on the multi-choice VQA samples of 52 tasks in MMIU. Formally, a task vector is defined by the weight variation between the weight fine-tuned on task data $D^t$ and the initial weight $W_0$ of a probing model with the minimum loss, as given by

\begin{equation}\label{eq:taskvec}
    V^t = \arg\min_W \mathcal{L}(W|D^t) - W_0 
\end{equation}

Notice that for most tasks, we train for 20 epochs, while for a subset of tasks with lower accuracy after initial training, we extend the training to 60 epochs to obtain accurate task vectors. Given the large parameter size of QwenVL-chat, we use LORA \citep{hu2021lora} for fine-tuning, which reduces the length of the task vector from 9.6B to 3.5M and consumes fewer storage resources.
2) For any two task vectors $V^s$ and $V^t$, we compute the cosine distance between their task vectors, where $G^{st} = 1 - \cos(V^s, V^t)$. This process results in a 52x52 task map.
3) As shown in Figure \ref{fig:taskmap} (b), the task map for MMIU reveals that similar tasks cluster together, such as GNAP and GuAR, which are related to GUI, and VO, TO, and VCo, which involve designing image sequence reasoning tasks, which show that the constructed task map aligns with intuition. A comprehensive breakdown of task map can be found in Table \ref{tab:taskmapcluster} and Table \ref{tab:long}.

\begin{xltabular}{\textwidth}{>{\hsize=0.4\hsize}l|>{\hsize=1.6\hsize}X|>{\hsize=0.4\hsize}X}
\caption{Details of task clustering on the task map of our MMT-Bench.} \label{tab:long} \\
\toprule \multicolumn{1}{c}{\textbf{Cluster ID}} & \multicolumn{1}{|c|}{\textbf{Tasks}} & \multicolumn{1}{c}{\textbf{\# Number}} \\ \midrule 
\endfirsthead

\multicolumn{3}{c}%
{\tablename\ \thetable{} -- continued from previous page} \\
\toprule \multicolumn{1}{c}{\textbf{Cluster ID}} & \multicolumn{1}{|c|}{\textbf{Tasks}} & \multicolumn{1}{c}{\textbf{\# Number}} \\ \midrule 
\endhead

\bottomrule
\endfoot

\bottomrule
\endlastfoot

1  & emotion-recognition,
forensic-detection,
visual-quality-assessment,
visually-grounded-reasoning,
visual-correspondence,
semantic-correspondence,
functional-correspondence, & 7 \\
\midrule
2 & spot-the-diff,
Multiples-image-captioning,
Homography-estimation,
single-object-tracking,
point-tracking,
jigsaw-puzzle-solving,
threeD-Pose-Estimation,
Image-Captioning-with-Spatial-Context,
Image-Spatial-Transformation-Estimation &  9\\
 \midrule
3 & next-img-prediction,
spot-the-similarity,
ravens-progressive-matrices,
threed-indoor-recognition,
threed-cad-recognition,
Multiview-reasoning,
threeD-Depth-Estimation & 7\\
 \midrule
4 & temporal-localization,
visual-cloze,
person-reid,
text2image-retrieval,
vehicle-retrieval,
face-retrieval,
sketch2image-retrieval,
image2image-retrieval,
handwritten-retrieval,
Icon-Question-Answering-with-Spatial-Context & 10\\
 \midrule
5 & mevis,
threeD-Scene-Reconstruction,
threeD-Object-Detection,
threeD-Object-Tracking & 4\\
\midrule
6 & visual-coherence,
visual-ordering,
temporal-ordering & 3 \\
\midrule
7 &meme-vedio-understanding,
action-quality-assessment,
video-captioning,
general-action-recognition,
textual-cloze,
casuality-reasoning,
Image-text-retrieval-with-Spatial-Context,
Egocentric-Video-QuestionAnswering & 8\\
\midrule
8 & gui-next-action-prediction,
gui-app-recognition,
Multiview-Action-Recognition,
threeD-question-answering & 4\\

 \bottomrule
\end{xltabular}

\subsection{Analysis}

We perform hierarchical clustering on the task map and analyze each cluster. Unlike the previous method of classification through multiple relationships, this approach leverages Qwen-VL as a probe network, allowing for a more objective segmentation based on the intrinsic attributes of the tasks themselves. Combining the model's performance in each cluster, as shown in Figure \ref{fig:taskmap} (b), with the task map presented in Figure \ref{fig:taskmap} (b), we begin by analyzing the in-domain tasks where the model demonstrates strong performance. Next, we examine the out-domain tasks where the model underperforms. Finally, we propose using Taskmap to assess task difficulty, guiding future model development. 

\begin{wraptable}{rt!}{0.5\textwidth}
\centering

\centering
\caption{The number of tasks within each cluster after hierarchical clustering, and the Spearman correlation $r$ between the average performance of the model on these tasks and the overall performance of the model.}
\scalebox{0.58}{
\begin{tabular}{lcccccccc}
\toprule
\# Cluster & 1 & 2 & 3 & 4 & 5 & 6 & 7 & 8  \\
\midrule
\# Tasks  & 7 & 9 & 7 & 10 & 4 & 3 & 8 & 4 \\
\midrule
\# $r$  & 0.94 & 0.85 & 0.92 & 0.96 & 0.83 & 0.93 & 0.74 & 0.73 \\
\midrule
\# Acc  & 27.9 & 34.7 & 26.3 & 26.8 & 17.2 & 20.2 & 32.3 & 33.7 \\
\bottomrule
\end{tabular}
}
\label{tab:taskmapcluster}

\end{wraptable}

\textbf{In-Domain Tasks Analysis. } In-domain tasks are tasks that most current multimodal large models can handle correctly. For multi-image tasks, the model generally struggles to achieve satisfactory results, with most models performing worse than random selection. Consequently, the model can only achieve good performance on a limited number of tasks. Specifically, as shown in Table \ref{tab:taskmapcluster}, for tasks in clusters 7, 8, and some tasks in cluster 2, which involve recognition or captioning (e.g., video captioning, action recognition), the model performs relatively well. We believe this is because these multi-image tasks focus on overall image perception, requiring less comparison and reasoning between images. Additionally, the model has already demonstrated strong capabilities in high-level perception tasks involving single images.

\textbf{Out-of-Domain Tasks Analysis.} Out-of-Domain Tasks refer to tasks where most models perform poorly. Specifically, as shown in Table \ref{tab:taskmapcluster}, we find that models struggle with tasks in clusters 4, 5, and 6. Upon analysis, we discover that tasks in clusters 4 and 6 involve modeling semantic relationships or sequential order among multiple images, which requires strong memory capabilities and advanced perceptual and reasoning skills. Most open-source models underperform on these tasks, especially in image sequencing problems (e.g., temporal ordering tasks), where even closed-source models struggle to achieve satisfactory results. Tasks in cluster 5 pertain to visual tasks involving 3D spatial relationships, such as detection and tracking. Although models show some proficiency in handling 2D visual tasks, they lack optimization for 3D data, making them capable of handling only simple 3D tasks that involve basic semantic understanding, but inadequate for accurately modeling complex 3D visual tasks.

\end{document}